\documentclass[letterpaper, 10 pt, conference]{ieeeconf}  % Comment this line out if you need a4paper

\IEEEoverridecommandlockouts                              % This command is only needed if 
\input{header}

\usepackage{newtxtext} % newtxmath
\usepackage{newtxmath} 
\usepackage{microtype}
\usepackage[utf8]{inputenc}
\usepackage[T1]{fontenc}
\usepackage{times} % assumes new font selection scheme installed

\title{ \LARGE \bf \emph{FruitNeRF++}: A Generalized Multi-Fruit Counting Method Utilizing Contrastive Learning and Neural Radiance Fields \\
}

\author{Lukas Meyer, Andrei-Timotei Ardelean, Tim Weyrich, and Marc Stamminger%
\thanks{The authors are with the Visual Computing Erlangen (VCE), Friedrich-Alexander-Universität Erlangen-Nürnberg-Fürth, Germany, E-Mail: {\tt\small [lukas.meyer, timotei.ardelean, marc.stamminger]@fau.de}}
\thanks{FruitNeRF Code: \href{https://github.com/meyerls/FruitNeRF}{github.com/meyerls/FruitNeRF}}
}

\begin{document}

%https://www.fau.de/research/service-fuer-forschende/publizieren-und-open-access/tipps-zur-korrekten-affiliation/

\twocolumn[{
\renewcommand\twocolumn[1][]{#1}
\maketitle
\includegraphics[width=0.33\linewidth]{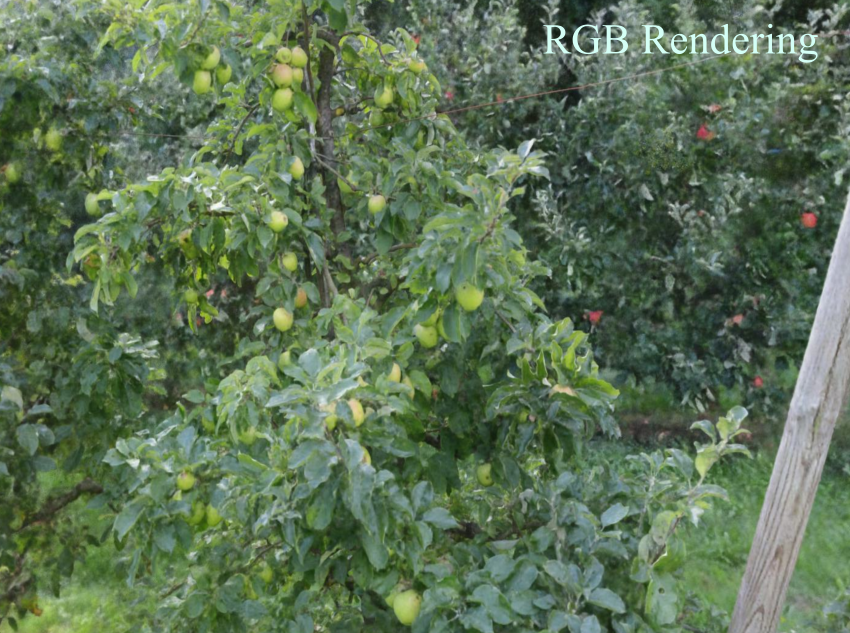}
\includegraphics[width=0.33\linewidth]{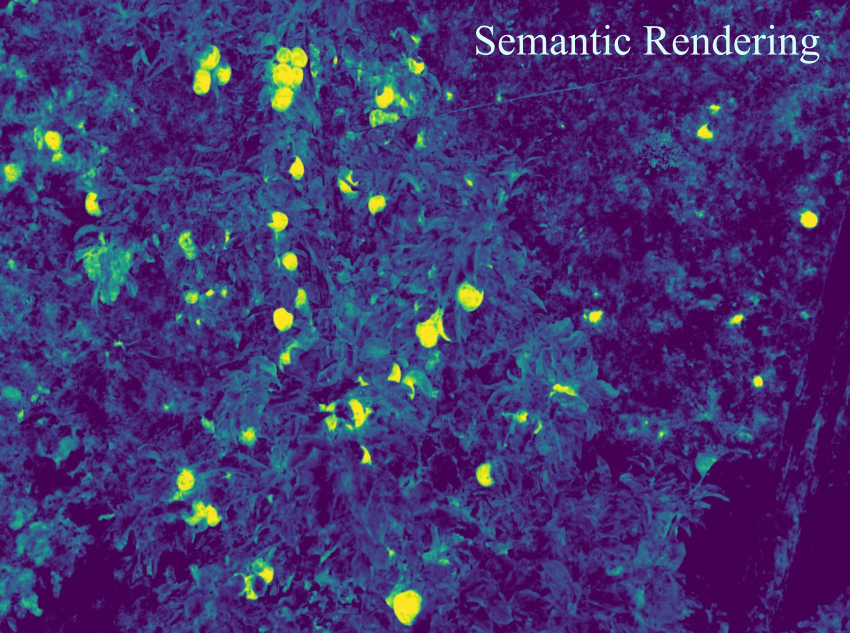}
\includegraphics[width=0.33\linewidth]{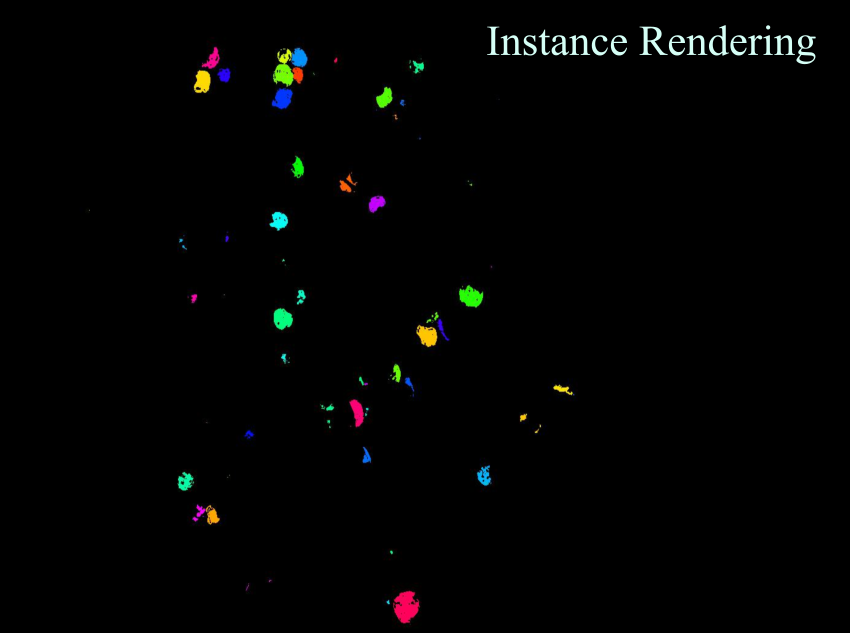}

\captionof{figure}{Rendering of RGB, semantic and instance images. For visualization of the results visit the project page: \href{https://meyerls.github.io/fruit_nerfpp}{meyerls.github.io/fruit\_nerfpp}.} 
%\vspace{1.55em}
\vspace{1em}
\label{fig:tile_figure}
}]
\thispagestyle{empty}
\pagestyle{empty}

\renewcommand{\thefootnote}{\fnsymbol{footnote}}
\footnotetext[1]{The authors are with the Visual Computing Erlangen (VCE), Friedrich-Alexander-Universität Erlangen-Nürnberg-Fürth, Germany, E-Mail: \texttt{[lukas.meyer, timotei.ardelean, tim.weyrich, marc.stamminger]@fau.de}}
\footnotetext[2]{This project is funded by the 5G innovation program of the German Federal Ministry for Digital and Transport under the funding code 165GU103B and the European Union’s Horizon 2020 research and innovation program under the Marie Skłodowska-Curie grant agreement No 956585.
The authors gratefully acknowledge the scientific support and HPC resources provided by the Erlangen National High Performance Computing Center (NHR@FAU) of the Friedrich-Alexander-Universität Erlangen-Nürnberg (FAU). 
The hardware is funded by the German Research Foundation (DFG).}

\renewcommand{\thefootnote}{\arabic{footnote}}
\setcounter{footnote}{0}

%%%%%%%%%%%%%%%%%%%%%%%%%%%%% Start Text  %%%%%%%%%%%%%%%%%%%%%%%%%%%%

%%%%%%%%%%%%%%%%%%%%%%%%%%%%% Abstract    %%%%%%%%%%%%%%%%%%%%%%%%%%%%

\begin{abstract}
We introduce \emph{FruitNeRF++}, a novel fruit-counting approach that combines contrastive learning with neural radiance fields to count fruits from unstructured input photographs of orchards. 
Our work is based on \emph{FruitNeRF}~\cite{FruitNeRF}, which employs a neural semantic field combined with a fruit-specific clustering approach.
The requirement for adaptation for each fruit type limits the applicability of the method, and makes it difficult to use in practice.
To lift this limitation, we design a shape-agnostic multi-fruit counting framework, that complements the RGB and semantic data with instance masks predicted by a vision foundation model.
The masks are used to encode the identity of each fruit as instance embeddings into a neural instance field.
By volumetrically sampling the neural fields, we extract a point cloud embedded with the instance features, which can be clustered in a fruit-agnostic manner to obtain the fruit count.
We evaluate our approach using a synthetic dataset containing apples, plums, lemons, pears, peaches, and mangoes, as well as a real-world benchmark apple dataset. 
Our results demonstrate that \emph{FruitNeRF++} is easier to control and compares favorably to other state-of-the-art methods.
%Project page: \url{https://meyerls.github.io/fruit_nerf++}

\end{abstract}

\section{Introduction}

Counting fruits and estimating yield is a valuable tool for effective harvest and post-harvest management \cite{fcsota}. 
These tasks include planning the harvest workforce, packaging and wrapping, as well as storage and processing. 
A versatile multi-fruit counting framework is essential to ensure applicability across different types of orchards.
Most existing frameworks are optimized for specific fruit types \cite{fuji_count, robust_fruit_counting, liu2019monocular, hanI2020comparative}, leaving a gap in solutions that can handle diverse fruit varieties.
To address this, we introduce a multi-fruit counting framework that unifies the detection process using a visual foundation model and generalizes object counting to accommodate varying shapes and appearances.

% In previous work, \emph{FruitNeRF}~\cite{FruitNeRF}, the first framework capable of simultaneously counting different types of fruit was introduced. 
\emph{FruitNeRF}~\cite{FruitNeRF} introduced a fruit-counting framework composed of two stages: one that is fruit-agnostic, and one that is fruit-specific.
The generic component leverages a neural semantic field to integrate RGB and semantic modalities in a 3D space. 
The second component generates a semantic point cloud from the NeRF, focused exclusively on fruit data, and applies a cascaded clustering technique to accurately count the fruits.
However, the cascaded point clustering method employed by \emph{FruitNeRF} presents a key limitation: it relies on fruit-specific shape priors, restricting its applicability to generalized multi-fruit counting. 
To address this limitation, we propose \emph{FruitNeRF++}, overcoming this issue by encoding instance-level information into a neural instance field using contrastive learning. 
This allows us to extract an instance-embedded point cloud from the neural field, which can be clustered in a shape-agnostic manner. 
Preliminary experiments demonstrate that this approach extends beyond fruit counting and is even applicable to more general object types.

%We identified the cascaded point clustering to be the main shortcoming
%\todo[inline]{From Tim: Start by listing the shortcoming that “FruitNeRF is only shown to work for apples (speculate on reasons, such as high contrast against leaves, fruit size relative to leaves, near-spherical shape, …) but, as we will show, does not easily extend to other types of fruit, even when re-trained.” Then, say that you will show(?) that you identified the clustering as the main hurdle, which prevents FruitNeRF from generalising.} 
%of \emph{FruitNeRF}, due to a fruit-specific clustering pipeline.
%Our work tackles this limitation by encoding instance-level information directly into a neural field using contrastive learning. 
%We then extract an instance-embedded point cloud from the neural field, which can simply be clustered in a shape-agnostic fashion. In fact, preliminary experiments show that our approach can even extend to more general object types.
% \todo[inline]{From Tim: first state suitability for multiple types of fruit, and then follow up with “In fact, as we will show, it even extends to more general object types.}
\if false
    Our pipeline consists of three main components, as depicted in Fig. \ref{fig:fruitnerfpp_Pipeline}.
    First, we compute instance masks for different fruit types using two foundation models: Grounded SAM \cite{ren2024grounded} and Detic \cite{detic}. 
    Using the posed images, we then train a neural radiance field, referred to as \emph{FruitNeRF++}, which consists of Appearance, Density, Semantic, and Instance Fields. 
    The Semantic Field encodes whether a point in space belongs to a fruit, while the Instance Field stores a $N$-dimensional embeddings that acts as identifier for each fruit.
    We then uniformly sample the implicit space to extract information about its density, semantics, and instances, which we store in a point cloud. 
    Finally, by clustering the instance fruit cloud, we can obtain an accurate fruit count.
\fi
The main contributions of our work are:

\begin{itemize}
    \item We propose a novel multi-fruit, counting method that is shape-agnostic, by employing a neural instance field. % instance neural radiance fields
    \item We extend a multi fruit dataset\footnote{Project website: \href{https://meyerls.github.io/fruit_nerfpp}{meyerls.github.io/fruit\_nerfpp}} \cite{FruitNeRF} with instance masks.
    \item We release the code of \emph{FruitNeRF++}\footnote{\emph{FruitNeRF++} code: \href{https://github.com/meyerls/FruitNeRFpp}{github.com/meyerls/FruitNeRFpp}} upon acceptance.
\end{itemize}
% \JO{One extra contribution: fruit agnosticity.}

\section{Related Work}
\subsection{Fruit Counting}
With the rise of computer vision in agriculture, fruit counting in various orchard environments has garnered attention across multiple disciplines, including precision agriculture. 
Most research focuses on fruit types such as sweet peppers~\cite{pagnerf}, strawberries~\cite{hqstrawberry, berryharvest}, apples~\cite{hanI2020comparative, robust_fruit_counting, fuji_count, ZHANG2022107062} and mangoes~\cite{liu2019monocular}.

Liu et al.~\cite{liu2019monocular} propose a fruit counting system that processes image sequences. During the detection phase, a convolutional neural network is used to detect mangoes. In the tracking phase, a Kalman Filter, paired with an optical flow tracker, is applied to track individual fruits over time. 
The final count is achieved by triangulating the tracked fruit instances and localizing them in 3D space.
Häni et al.~\cite{hanI2020comparative} utilize a U-Net architecture \cite{unet} to predict segmentation masks for apples. 
These masks are projected into 3D to form a semantic point cloud. 
The fruit count is then obtained by clustering the point cloud and projecting the resulting clusters back onto the image plane.
Similarly to Häni et al.~\cite{hanI2020comparative} the approach of Gené-Mola et al.~\cite{fuji_count} uses instance segmentation masks. 
By clustering the semantic point cloud and projecting it back onto the image plane, they assign each 3D cluster a fruit id through overlap with the instance masks.
\emph{FruitNeRF}~\cite{FruitNeRF} is the first approach to work on different fruit types leveraging neural radiance fields and visual foundation models.
By lifting the RGB and semantic images to 3D, a semantic point cloud is extracted.
Clustering the semantic point cloud combined with fruit-specific templates leads to a fruit count.

\begin{figure*}[ht]
    \centering
    \includegraphics[width=1\linewidth]{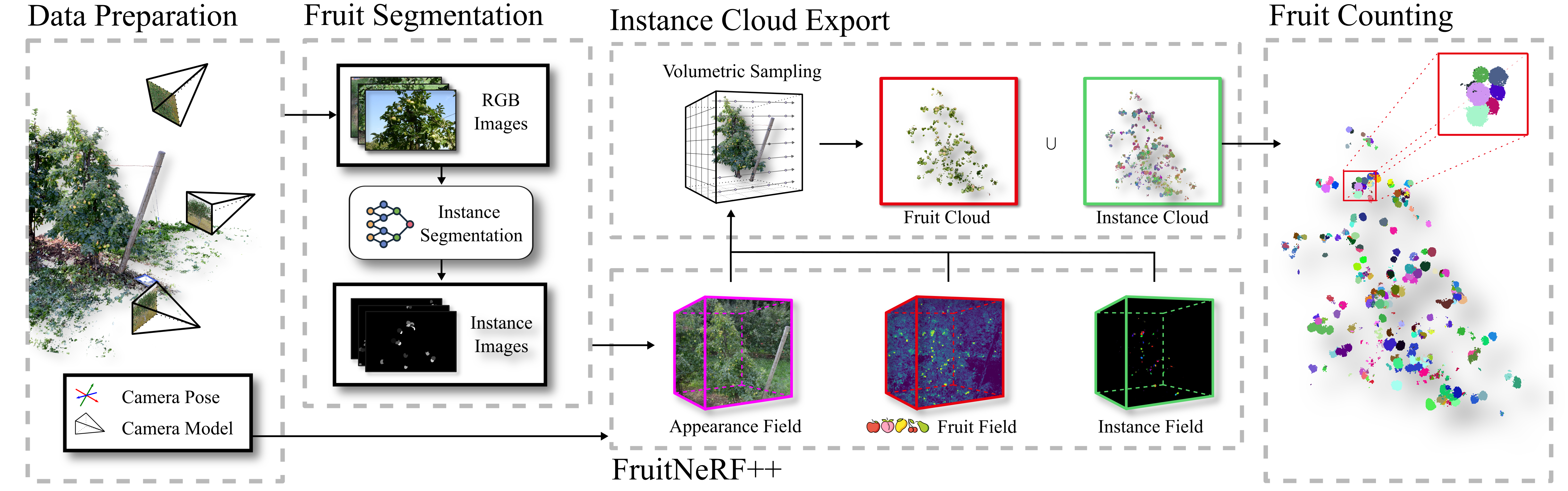}
\caption[]{Pipeline of \emph{FruitNeRF++}. For the images we recover both intrinsic and extrinsic camera parameters. We then extract semantic and instance masks for arbitrary fruit types using SAM \cite{kirillov2023segany} and Detic \cite{detic}. The data are used to train a neural radiance field with an neural appearance, semantic (fruit) and instance field. By clustering the combination of a fruit and instance point cloud we obtain a precise fruit count.}
\label{fig:fruitnerfpp_Pipeline}
\end{figure*}

\subsection{Contrastive Learning}

% \todo[inline]{@Timotei: [WIP]}
One of the prominent approaches for self-supervised learning of representations and dimensionality reduction is to employ a constrastive objective~\cite{hadsell2006dimensionality}. 
The essence of this class of methods is to use positive (similar) and negative (dissimilar) pairs of items as supervision.
In general, these are significantly easier to obtain compared to exhaustive labels, and can sometimes be derived from simple priors, such as image augmentations when learning visual representations~\cite{simclr}. 
Thanks to the versatility of the concept, contrastive learning has been used in various contexts such as: unsupervised image classification~\cite{simclr}, semantic segmentation~\cite{stego}, anomaly detection~\cite{cadet}, anomaly clustering~\cite{blindal}, or learning audio representations~\cite{cola}.
The various methods adopt different contrastive loss functions; some of the most common include the hard margin contrastive loss~\cite{hadsell2006dimensionality}, triplet loss~\cite{tripletloss}, ArcFace~\cite{arcface}, and NT-Xent~\cite{ntxent}.

Our method adapts the InfoNCE loss~\cite{infonce} to leverage multiple positive and negative pairs, which are mined from 2D instance segmentation masks. The contrastive objective guides the training of our neural instance field through volumetric rendering, yielding a 3D-consistent instance field. 
Most related to our work is the use of contrastive learning for optimizing a 3D representation~\cite{contrastivelift}, discussed in the following subsection.
%; this method is described in the following subsection.

\subsection{Neural Panoptic Fields}
Since the introduction of neural radiance fields (NeRFs)~\cite{nerf} numerous advances have been made mainly focusing on the improvement of quality~\cite{zipnerf}, inference~\cite{kilonerf, plenoctrees} and training time~\cite{plenoxels, instantngp}. 
Additionally NeRFs have been extended by both a semantic~\cite{zhi2021inplace} and instance~\cite{instancenerf} field component. 
\emph{SemanticNeRF}~\cite{zhi2021inplace} extends the default architecture by a semantic field to jointly optimize for semantics with rgb and density. 
Thus, they obtain a multi-view-consistent rendering of the assigned semantic classes.
\emph{InstanceNeRF}~\cite{instancenerf} uses instance masks that are already view-consistent; this enables the instance id of every object to be learned directly through an underlying instance field.
% \emph{InstanceNeRF} \cite{instancenerf} on the other hand uses images with labeled multi view-consistent instance masks and encodes for every object an instance id to an underlying instance field.

Panoptic Lifting~\cite{panopticlifting} and Contrastive Lift~\cite{contrastivelift} address a similar problem to ours, in that the methods seek to reconstruct a 3D panoptic/instance field from 2D masks. 
In Panoptic Lifting, the predicted 2D instances are not 3D-consistent; to solve this inconsistency, the method computes injective mappings from 2D instances to 3D surrogate identifiers by solving linear assignment problems. 
%The 2D instances are predicted independently from input images, using a pretrained neural network, yielding instance masks that are not 3D-consistent.
%To attend to this inconsistency, Panoptic Lifting computes injective mappings from 2D instances to 3D surrogate identifiers by solving linear assignment problems. 
However, the mappings have to be computed repeatedly, imposing a large computational cost.
Contrastive Lift avoids an explicit mapping by employing a constrastive objective to learn a field of instance embeddings. 
%This is optimized by rendering the embeddings and using the 2D instance masks to bring closer the features corresponding to the same instance and separate the embeddings from different instances. 
Here, the authors render the feature embeddings and utilize the 2D instance masks to coalesce similar and penalize dissimilar features.
%\JO{Here, the authors render the feature embeddings and utilize the 2D instance masks to coalesce similar and penalize dissimilar features.}
%To improve the robustness of the method, the authors introduce an additional \emph{concentration} loss and use a \emph{slow} teacher that is updated as an exponential moving average of the instance field.
To improve the robustness of the method, the authors introduce an additional \emph{concentration} loss and use a \emph{slow} teacher for training the instance field.
%\JO{their method. Just out of curiosity: Why does it make the method more robust?}
We also use contrastive learning to optimize a field of instance embeddings; however, we use a different loss function that avoids the overhead of maintaining two different (teacher and student) networks. 
Moreover, our approach is tailored for small objects, which are generally difficult to model. 
This is done by adjusting the pixel sampling (Sec. ~\ref{ssec:pixelsampler}) to focus on the pairs that are hard to distinguish.
% \JO{Either leave out the last subordinate clause (Nebensatz), or go more into detail on \_how\_ the sampling is adjusted in your case. }

 \if false
\section{Fundamentals}
%\todo[inline]{From Tim: Before 'Methodology', would it make sense to introduce FruitNeRF and analyse its shortcomings? The section could be called 'FruitNeRF'. Shortcomings could be listed with forward-pointers to results.}

\todo[inline]{I think we can simply remove this section, and leave just preliminaries with some opening text like: As our method builds on FruitNeRF, we summarize here the relevant components. At a high-level, FruitNeRF consists of a 3D representation that models the geometry and semantics (fruit vs background), and an elaborate clustering algorithm that identifies individual fruits. To represent the geometry, a NeRF is used, which is a neural representation that implicitly...}

Classic fruit-counting methods use either image-based counting methods or 3D point clouds from fruit trees to count fruits. 
Compared to image-based methods, which rely on fruit specific-features, 3D reconstructions have the advantage to project semantic or instance information into space and count the fruits in 3D.
The natural development is to generalize the fruit detection process for different fruit types and secondly change the underlying representation.
Therefore we proposed in our previous work, \emph{FruitNeRF}, to use a visual foundation models such as SAM \cite{kirillov2023segany} for instance segmentation and neural radiance fields as a new representation type.

\fi

%In our previous work we proposed\emph{FruitNeRF}.
%A fruit counting framework that's key component was build up on NeRFs.

%Our pipeline extends \emph{FruitNeRF} by adding an instance field alongside the color and semantic fields. 
%In this section we discuss the different neural fields we employed and their rendering pipelines.

% NeRFs \cite{nerf} utilize a continuous function $\mathcal{F}$ to describe a scene which is encoded in a implicit neural representation. 
\section{Preliminaries}
As our method builds on \emph{FruitNeRF}~\cite{FruitNeRF}, we summarize here the relevant components. 
\emph{FruitNeRF} consists of a 3D representation that models the geometry and semantics, and an elaborate clustering algorithm that identifies individual fruits. 
To represent the geometry, a NeRF \cite{nerf} is used, which is a neural representation that implicitly encodes a scene through a continuous function $\mathcal{F}$.
Given a set of images and their camera parameters, a neural network is optimized to learn geometric and appearance properties of the scene.
This is achieved by optimizing a density field $\mathcal{F}_{\sigma}\!\!: \vect{x} \rightarrow \sigma$ that maps a spatial coordinate $\vect{x} \in \mathbb{R}^{3}$ to a volume density $\sigma$.
To obtain the RGB radiance $\vect{c}$, the appearance field $\mathcal{F}_{c}\!\!: (\vect{x}, \vect{d}) \rightarrow \vect{c}$ takes the position $\vect{x}$ and view direction $\vect{d}$ as input.
% To determine the color composition $\mathbf{c}$ of a pixel we march 
The composed color of a pixel is determined by marching along a camera ray $\vect{r}(t) = \vect{o}+t\vect{d}$ that originates in the camera center $\vect{o}$.
% and $\vect{x}$ to the selected pixel position. 
The estimated color $\hat{\vect{C}}$ is obtained through volumetric rendering by querying $\mathcal{F}$ in $K$ points along the ray $\vect{r}$ and accumulating the color values based on the density:
% By querying $\mathcal{F}$ with $K$ points along the ray $\mathbf{r}$ we are able to accumulate the density and color values to obtain the estimates color $\hat{\mathbf{C}}$ using volumetric rendering:
\begin{equation}
\begin{split}
 & \vect{\hat{C}}(\vect{r}) = \sum_{k=1}^K \hat{T}(t_k) \alpha(\sigma(t_k)\delta_k)\vect{c}(t_k), \\
 \text{ where } \hspace{0.5em} & \hat{T}(t_k) = \exp\left(-\sum_{a=1}^{k-1}\sigma(t_a)\delta_a\right)\;\;.
 \end{split}
\end{equation}
\noindent Here $\delta_k = t_{k+1} - t_k$ defines the distance between two adjacent points and the transmittance probability is given by $\alpha(x) = 1 - \text{exp}(-x)$. An illustrative RGB rendering of a fruit tree is visualized in Fig. \ref{fig:tile_figure} on the left.

Additionally, \emph{FruitNeRF}~\cite{FruitNeRF} extends the NeRF approach by a neural semantic field~\cite{zhi2021inplace} to lift 2D semantic information into 3D.
Along with the color, a semantic field $\mathcal{F}_{s}\!\!: \vect{x} \rightarrow s \in \mathbb{R}$ is introduced, which maps a spatial coordinate to a semantic value. 
The estimation of the semantic $\vect{\hat{S}}$ for a given pixel is computed similarly to $\vect{\hat{C}}(\mathbf{r})$, but instead of colors $\vect{c}(t_k)$, semantics $\vect{s}(t_k)$ are accumulated according to
\begin{equation}
\begin{split}
 & \vect{\hat{S}}(\vect{r}) = \sum_{k=1}^K \hat{T}(t_k) \alpha(\sigma(t_k)\delta_k)s(t_k)\;\;.\\
 \end{split}
\end{equation}

\noindent The semantics rendering is depicted in Fig. \ref{fig:tile_figure} in the middle.
By volumetric sampling of the implicit neural fields a point cloud is obtained that contains semantic points with a high density. Such semantic point cloud is referred to as a fruit point cloud. %  also denoted as fruit point cloud.
%By volumetric sampling of the implicit neural fields we obtain a point cloud which contains only points that have a high density whether it belongs to the trunk, foliage, ground, or fruit.
%If we masks all points by their semantics we create a fruit points cloud, which only contain dense points associated to a fruit.
%\JO{These two sentences are kinda weird.}
%\JO{By rvolumetric sampling the implicit neural fields, we obtain a point cloud that contains points with a high density in regions of semantic value (?), i.e. the density is high when a point belongs to the trunk, foliage, ground or fruit.}
%\JO{We can then create a point cloud of fruits by masking out points with low semantic value.}
%\JO{Irgendwie sowas.}
Afterwards, a cascaded clustering approach is applied in \emph{FruitNeRF}, isolating each fruit to obtain a fruit count.
%As the clustering is performed just in Euclidean space, this approach has problems to differentiate between conglomerated fruits.
%\JO{As the clustering is only able to separate fruits in euclidean space [...]}
\emph{FruitNeRF} uses a fruit-specific template matching approach to subdivide clusters.
The limitation of the clustering approach is that it needs fruit-templates for different type of fruit and varying template sizes for different grow stages.

To avoid this need for fruit-specific templates, we introduce with \emph{FruitNeRF++} the concept of neural instance fields to not only learn semantics but also an instance identifier for each fruit in the scene.
This allows us to combine semantic and instance information in the clustering stage and forego the template matching.
%\JO{I just learned that the thing I thought meant illicit, is actually written as elicit. And illicit means illegal. Guess I've written some illegal things in my papers. Oh well...}

\section{FruitNeRF++}

In this chapter we introduce the individual components of the \emph{FruitNeRF++} pipeline, which is visualized in Fig.~\ref{fig:fruitnerfpp_Pipeline}.

\subsection{Data Preparation}
\label{sec:datapreparation}
Both our synthetic and real-world datasets consist of sets of unordered RGB images. 
The synthetic dataset incorporates the models from \emph{FruitNeRF}, consisting of different fruit trees \cite{fruitassets}: apple, plum, lemon, pear, peach, and mango. 
For each tree model, we render 300 images from the upper hemisphere and extract the extrinsic and intrinsic camera parameters using the BlenderNeRF plugin \cite{BlenderNeRF}.
In addition to rendering color and semantic images, we extend the dataset by additionally rendering the ground-truth instance masks of the fruits.
A visualization of the dataset is shown on the project website.

The FUJI \cite{fujiapple} dataset consists of $582$ images taken from 12~trees at a commercial apple orchard. 
To recover the camera poses, we applied COLMAP \cite{schoenberger2016sfm} to each side and manually registered both sides. The semantic and instance masks were generated using the visual foundation models \emph{Grounded-SAM} \cite{ren2024grounded, liu2023grounding, kirillov2023segany} and Detic \cite{detic}.
%Additionally we used the instance segmentation masks from the original paper \cite{fujiapple}.
All images for training were downsampled to a resolution of $1296$\,px $\times$ $864$\,px.

\subsection{Fruit Segmentation}
\label{ssec:fruit_segmentation}

To achieve a unified fruit counting approach we utilize \emph{Grounded-SAM}~\cite{ren2024grounded} and Detic~\cite{detic} to predict precise instance segmentation masks without training a custom model.
Both foundation models have the capability to generalize well to different types of fruit without the need for fine-tuning.

\emph{Grounded-SAM} uses the name of each fruit as a text prompt, except for the fruit mango and plum, where we added \texttt{apple} to the prompt due to poor masks with \texttt{mango} or \texttt{plum} only.
Grounded-SAM \cite{ren2024grounded} combines the open-set object detector Grounding DINO \cite{liu2023grounding} with the open-world segmentation model SAM \cite{kirillov2023segany}.
For SAM \cite{kirillov2023segany} we use the weights of SAM-HQ \cite{samhq}.

Grounding DINO generates precise bounding boxes for each image by leveraging textual information as an input condition. 
These bounding boxes are then used by SAM as box prompts to predict accurate instance masks.

As Detic~\cite{detic} can detect every class available in the vocabulary collection LVIS (Large Vocabulary Instance Segmentation)~\cite{lvis}
Detic \cite{detic} combines both the identification of bounding boxes and the assignment of categories to objects. 
For Detic we choose the default network configuration and selected LVIS (Large Vocabulary Instance Segmentation) \cite{lvis} as a vocabulary collection.
For each detected bounding box, Detic computes a CLIP~\cite{CLIP} embedding vector and uses it to estimate a corresponding text label. 
By leveraging CLIP embeddings, Detic can predict object categories without having seen them during training. 
we select every fruit-related classes available in LVIS and filter the predictions in a post-processing step.

On average the masks predicted of \emph{Grounded-SAM} and Detic over all synthetic data have an intersection over Union (IoU) of $0.561$ and $0.49$, respectively. A detailed overview for different fruits is listed in Tab.~\ref{tab:segmentationresult}.
However, both approaches face the issue that when a fruit is partially occluded by an obstacle, such as a leaf or branch, it may be detected as two separate instances. 
This leads to the feature space of a fruit being partially separated and incorrectly counted as two separate instances in the clustering phase.

\begin{table}[t!]
    \definecolor{cellgreen}{RGB}{247,203,153}
    \centering
    \caption{Segmentation evaluation of Grounded-SAM~\cite{ren2024grounded} and Detic~\cite{detic} on synthetic data using Intersection over Union (IoU). For Grounded-SAM we used each fruit as a text prompt. For Detic we took an array of multiple fruit classes present in the LVIS~\cite{lvis} Vocabulary and applied it to all fruit images. Listed tags: \texttt{apple, apricot orange\_(fruit), peach, persimmon, mandarin\_orange, pear, banana, mango, lemon, pumpkin, plum, grape, cherry, blackberry, fig\_(fruit), blueberry, pinecone, raspberry, date\_(fruit), almond, lime, clementine}. The best performing segmentation model is highlighted in green.}
    \label{tab:segmentationresult}

    \begin{tabular}{l | c c | c c}
       IoU ($\uparrow$)  &  Grounded-SAM      &  Prompt        & Detic   & Classes   \\ \toprule
     apple &  \cellcolor{cellgreen} 0.659    &  \texttt{apple}         & 0.632   & \multirowcell{6}{see\\ caption}  \\
     pear  & \cellcolor{cellgreen}  0.519    &  \texttt{pear}          & 0.468   &    \\
     plum  & \cellcolor{cellgreen}  0.382    &  \texttt{plum, apple}         & 0.144   &    \\
     lemon & \cellcolor{cellgreen}  0.556    &  \texttt{lemon}         & 0.514   &    \\
     peach & \cellcolor{cellgreen}  0.662    &  \texttt{peach}         & 0.654   &    \\
     mango & \cellcolor{cellgreen}  0.588    &  \texttt{mango, apple} & 0.528   &    \\
    
    \end{tabular}

\end{table}

\subsection{Neural Instance Field}
\label{sec:cf_nerf}
%In this section we describe the key component of \emph{FruitNeRF++}, the neural instance field (Sec. \ref{ssub:instance_rendering}). 
%Along with the neural field we propose a constrastive loss function (Sec. \ref{ssub:lossfunction}) and an pixel sampler strategy (Sec. \ref{ssec:pixelsampler}) that prepares data for the training process. 
%Lastly, we detail the training process (\ref{ssec:cascaded}).
In this section we describe the key components of \emph{FruitNeRF++}: the neural instance field (Sec. \ref{ssub:instance_rendering}), the proposed contrastive loss function (Sec. \ref{ssub:lossfunction}), the pixel sampler strategy (Sec. \ref{ssec:pixelsampler}), and the training process (\ref{ssec:cascaded}).

\subsubsection{Instance Rendering}
\label{ssub:instance_rendering}

Similarly to Contrastive Lift~\cite{contrastivelift}, the aim of the instance field is to lift view-inconsistent 2D instance masks to 3D.
Therefore, we optimize the 3D instance embeddings through contrastive learning using 2D instance masks predicted by a foundation visual model.
Our approach makes important modifications targeted to fruit-counting, by improving the separation of small objects, which is described in the next section.
%in Sec. \ref{sec:implementationandtraining}.

The instance field $\mathcal{F}_{i}\!\!: \vect{x} \rightarrow \vect{i}$ maps 3D coordinates $\vect{x}$ to feature embeddings $\vect{i} \in \mathbb{R}^D$. 
The goal of the instance field is to map points that belong to the same fruit to similar embeddings, while keeping embeddings from different fruits separated. Thus the instance embedding acts as a fruit identifier.
% This feature embedding holds in a $D$-dimensional embedding information to which fruit id a point belongs. 
% To fit the instance ids of a fruit into the instance field we use the differentiable rendering equation,
The rendering of the instance field uses the same approach as Bhalgat et al. \cite{contrastivelift}:
\begin{equation}
\begin{split}
  \vect{\hat{I}}(\vect{r}) = \sum_{k=1}^K \hat{T}(t_k) \alpha(\sigma(t_k)\delta_k)\vect{i}(t_k)\;\;.
 \end{split}
\end{equation}
% to estimate the per-pixel embedding vector $\hat{\mathbf{I}}$. 
An instance rendering after clustering (see more regarding our clustering approach in Sec. \ref{sec:counting_instance_vector}) is visualized in Fig. \ref{fig:tile_figure} on the right.

\subsubsection{Contrastive Loss Function}
\label{ssub:lossfunction}

\begin{figure}[b]
    \centering
    \includegraphics[width=1\linewidth]{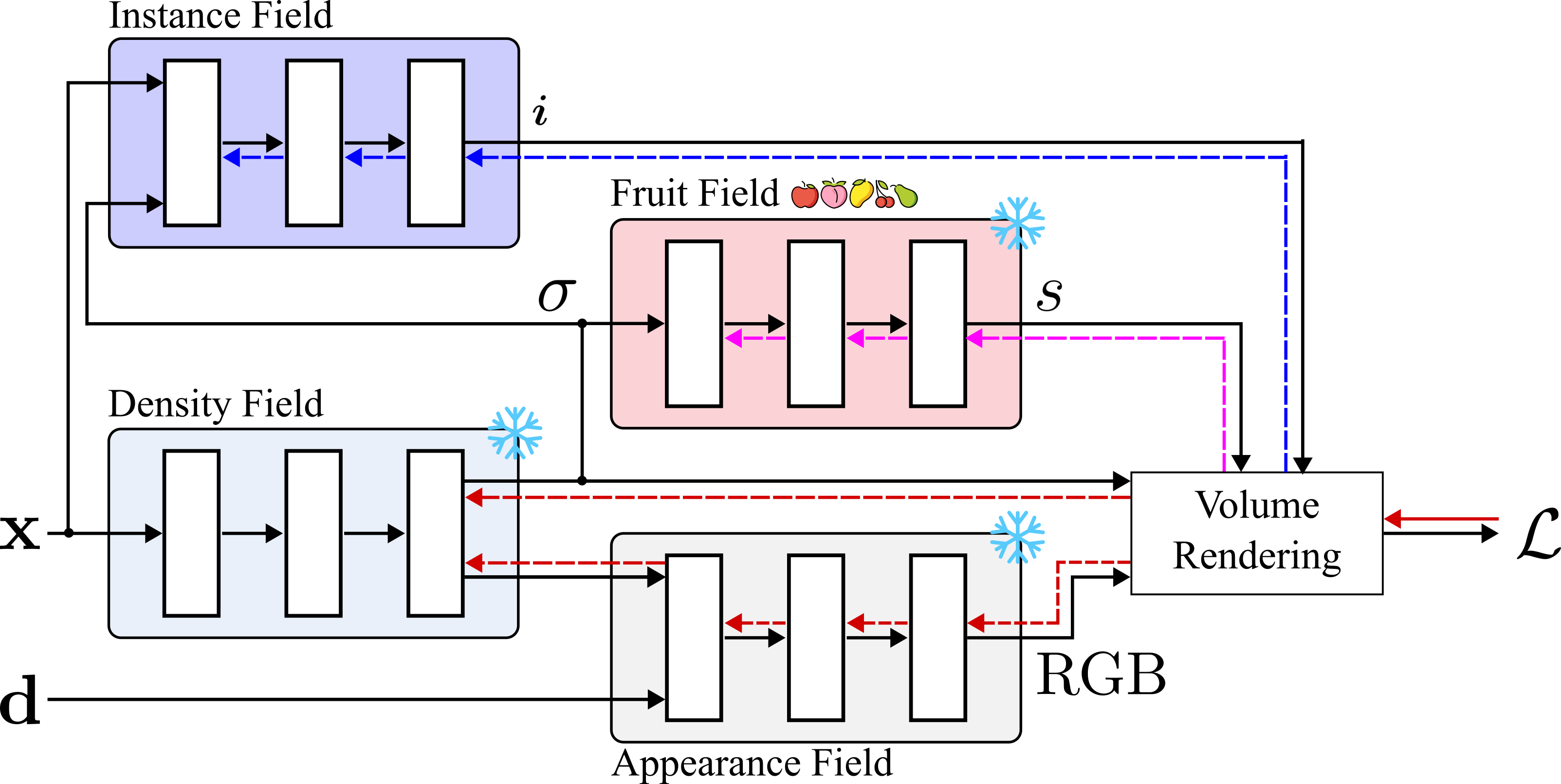}
    \caption{Overview of the \emph{FruitNeRF++} architecture, split up into four different components: density field, appearance field, fruit field, and instance field. %The components are detailed in~\ref{ssec:architecture}.
    The density field encodes the volume density $\sigma$, the appearance field the color $\textit{RGB}$, the Fruit Field the semantic information about the fruit in space, and the instance field a feature vector $\vect{i}$ encoding information about the instance group of a point in space. 
    The dashed arrow indicates the flow direction of the gradient. For training the different fields we employed a cascaded training scheme. 
    First we train the density and $\textit{RGB}$ alone, followed by activating the semantic Fruit Field. Lastly, we freeze all three neural fields and train only the instance field. 
    %More information on training can be found in Sec. \ref{sec:implementationandtraining}. 
    The figure is adapted from Özer et al. \cite{thermalnerf}. The colors of the dashed arrows correspond to their individual loss function in Eq.~\ref{eq:loss_func}}
    \label{fig:cf_nerf_architecture}

\end{figure}

We extend the base network by a semantic (also referred as Fruit) and instance field. 
An overview of the \emph{FruitNeRF++} architecture is schematically illustrated in Fig. \ref{fig:cf_nerf_architecture}. 
For training the density and appearance field we use the original photometric loss proposed by Mildenhall et al. \cite{nerf}.
It computes the loss between the ground truth pixel's RGB value $\vect{C}(\vect{r})$ and the predicted color value $\vect{\hat{C}}(\vect{r})$ along each ray $\mathbf{r}$ as
\begin{equation}
	\mathcal{L}_{\text{Photo}} = \frac{1}{|\mathcal{R}|}\sum_{\vect{r} \in \mathcal{R}}||\vect{C}(\vect{r}) - \vect{\hat{C}}(\vect{r})||_2^2\;.
\end{equation}
Here $\mathcal{R}$ is defined as the set of all rays.
For optimizing the semantic field, the loss has to distinguish between the two classes: background and fruit pixels. 
Therefore, we employ a binary cross entropy loss:
\begin{equation}
	\mathcal{L}_{\text{Sem}}\!=\!\frac{1}{|\mathcal{R}|}\sum_{\vect{r} \in \mathcal{R}}p(\vect{r})\log \hat{p}(\vect{r}) + (1- p(\vect{r}))\log (1-\hat{p}(\vect{r})).
\end{equation}

\noindent To train the instance field, we adapt the InfoNCE loss from MoCo~\cite{moco}:
\begin{equation}
        \mathcal{L}_q =-\log \frac{\exp(\langle \vect{q}, \vect{q_{+}}\rangle / \tau)}{\sum_{\vect{q_{-}} \in Q} \exp(\langle \vect{q}, \vect{q_{-}}\rangle / \tau) + \exp(\langle \vect{q}, \vect{q_{+}}\rangle / \tau)} \;\;,
\end{equation}

\noindent where $\tau$ represents the temperature of the softmax, $\vect{q_{+}}$ represents a positive pair with $\vect{q}$, and the denominator sums over all negative samples ($\vect{q_{-}}$) and one positive sample, $\vect{q_{+}}$.

In contrast, we have readily available multiple positive pairs for each sampled pixel, i.e., all pixels belonging to the same 2D instance. 
% Similarly, a negative pair can be formed with each pixel from all other fruits.
We find that a naive implementation, where each pair of pixels in one fruit is contrasted against all pixels from all other fruits, is suboptimal.
Instead, we compute a fruit-prototype feature vector $\vect{F}_a$, as the average of the features selected for that fruit: $\vect{F}_a = \frac{1}{P}\sum_{\mathcal{P}(x_j) = a}\vect{i}(x_j)$, where $\mathcal{P}(\vect{x})$ is the fruit to which a point $\vect{x}$ belongs.
For each pixel, we treat the fruit-prototype feature as the positive pair, and the prototypes of all other fruits as negative pairs. Thus, our contrastive loss is:
\begin{equation}
        \mathcal{L}_\text{contr} =-\!\!\!\!\!\sum_{j=1}^{G\times L \times P}\!\!\! \!  \log \frac{\exp(\langle \vect{i}(x_j), F_{\mathcal{P}(x_j)} \rangle / \tau)}{\sum_{a=1}^{G\times(L+1)} \exp(\langle \vect{i}(x_j), F_a \rangle / \tau)} \;\;.
\label{eq:contr}
\end{equation}
% \JO{Dude, wtf.}
\noindent where $G$, $L$, and $P$ represent the number of groups, the number of fruits per group, and the number of pixels per fruit respectively; sampled as discussed in \ref{ssec:pixelsampler}. 
The final training loss is finally given by:
\begin{equation}
	\mathcal{L} = \mathunderline{photo}{\mathcal{L}_{\text{Photo}}} + \mathunderline{sem}{\lambda_{\text{Sem}}\mathcal{L}_{\text{Sem}}} + \mathunderline{instance}{\lambda_{\text{contr}}\mathcal{L}_\text{contr}} \;\;.
    \label{eq:loss_func}
\end{equation}
The weighting factors $\lambda_{\text{Sem}}$ and $\lambda_{\text{contr}}$ are set to $1$, and we prevent both gradients to propagate trough the density field.

\subsubsection{Pixel Sampling Strategies}
\label{ssec:pixelsampler}

% Training \emph{FruitNeRF++} requires staggered training for each individual neural field. 
% For the density and appearance fields, we use an RGB pixel sampler, while the semantic field uses a semantic pixel sampler, and the instance field uses an instance pixel sampler. These pixel samplers differ in how they select pixels from one or multiple images in a training batch.

% In order to make the training process more stable, we optimize the instance field using a different sampling strategy compared to color and semantics. 
We observe that the uniform pixel sampling, customarily employed when training radiance fields, is inefficient in providing the training signal for the instance fields. 
That is because the large number of background pixels overshadows the small gradient updates needed to distinguish between different fruits.
Therefore, we design specific pixel samplers for training different fields, tailoring the optimization process for the fruit-centric tasks.

The purpose of the RGB and semantics pixel sampler is to provide uniform coverage of all data to train the density, appearance (RGB) and semantic fields. 
Therefore, we keep here the default method used in Nerfstudio~\cite{nerfstudio}, which uniformly samples pixels across all available images.
% It randomly samples pixels across all available images, which is the default method used in Nerfstudio \cite{nerfstudio}.

% \todo[inline]{@Timotei: Continue rewriting to be more succint.}

% Before using the instance masks for training the positional relationships between the fruits on a per image level have to be stored.
% This is archived by firstly computing the median pixel location of all available fruits.
% Afterwards, the euclidean distance between all fruit centers are computed.
% By sorting it by the nearest neighbour we are able to computed local fruit clusters which is relevant for sampling instance pixels according to their local and global neighbours. Image pixels are discussed in the following section.

\begin{figure}[t!]
     \centering
     \captionsetup[subfigure]{justification=centering}
     \begin{subfigure}[b]{0.492\linewidth}
         \centering
         \includegraphics[width=\textwidth]{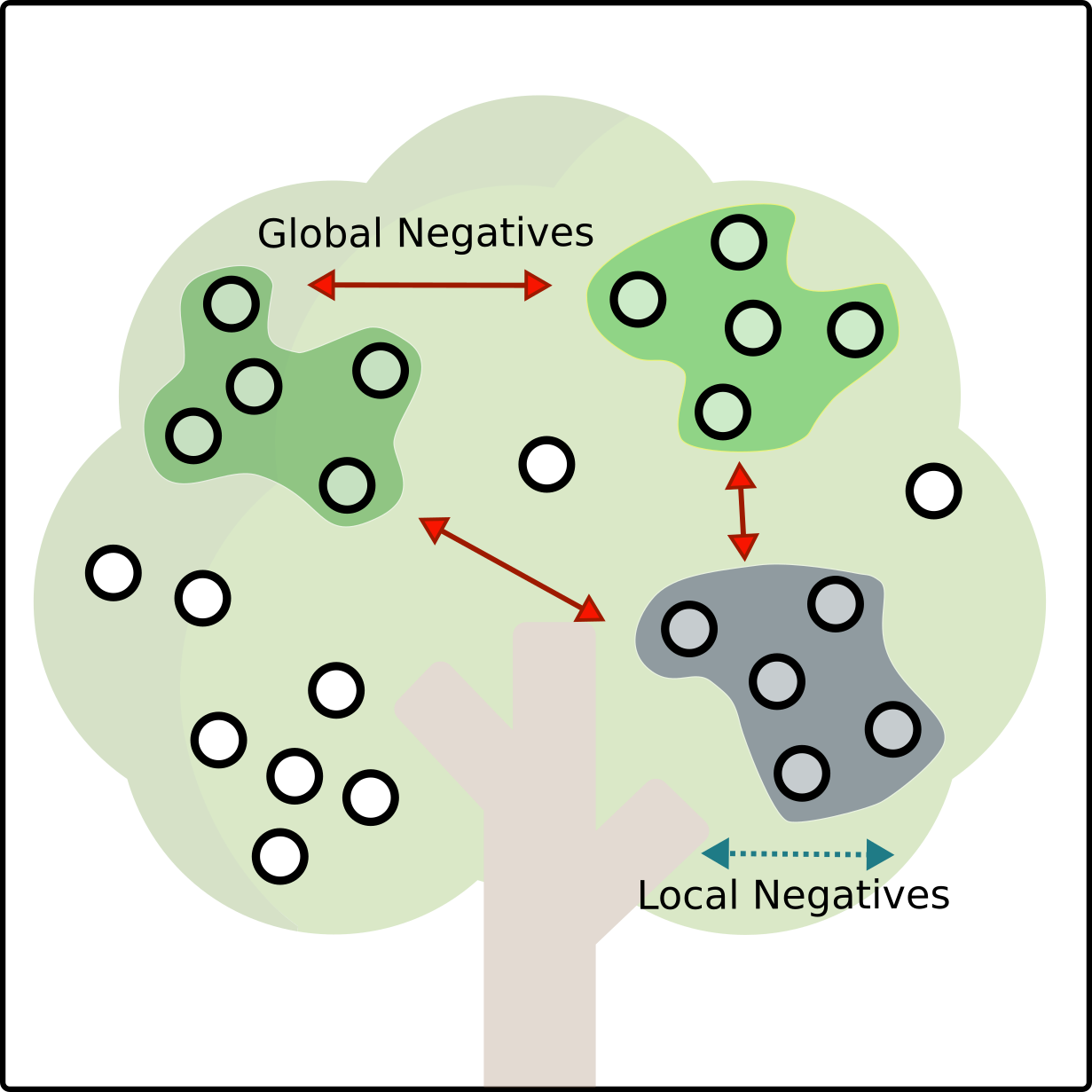}
         \caption{Concept of Global and \\ Local Negatives}
         \label{subfig:a}
     \end{subfigure}
     \hfill
     \begin{subfigure}[b]{0.492\linewidth}
         \centering
         \includegraphics[width=\textwidth]{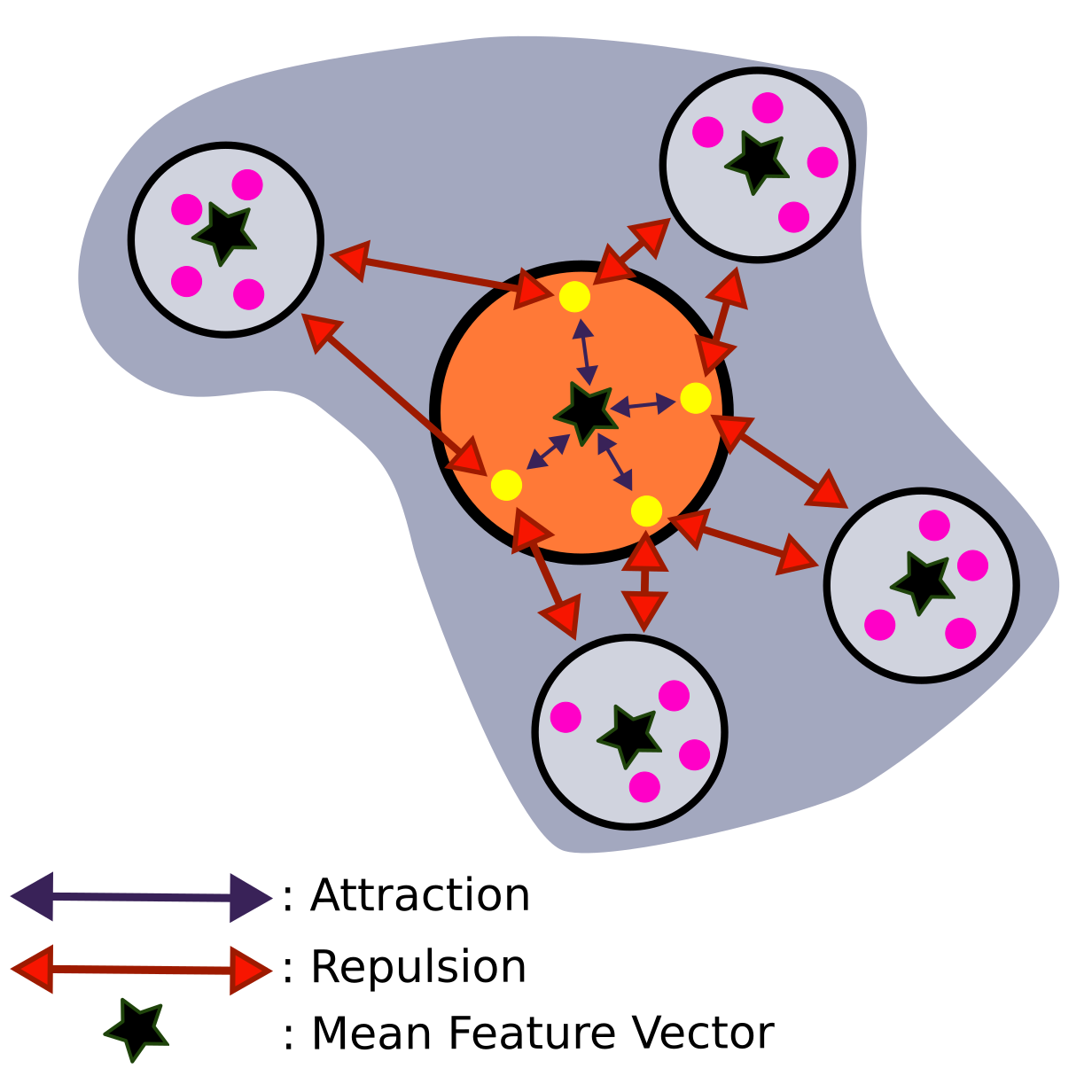}
         \caption{Illustration of \\ Group Pixel Sampling}
         \label{subfig:b}
     \end{subfigure}
\caption[]{In Fig. (a) we visualize the concept of local and global negatives. Local negatives are a collection of multiple fruits in near vicinity (see Sec. \ref{ssec:pixelsampler}). By selecting these hard negatives, we enforce the features of neighbouring fruits to be distinct, facilitating their separation during the clustering stage. Global (weak) negatives are then used to separate distant fruits. Fig (b) visualizes both pixel sampling and local negatives in detail. 
%For every group of fruits we sample a certain amount of pixels. 
The pixels from the orange center are denoted as positive and all others as negative. %By accumulating the instance features along a ray we compute for every selected pixel an $D$-dimensional feature vector and the normalized mean feature vector (denoted as a star) for every fruit. 
By computing the contrastive loss function between every pixel's feature vector and the mean feature vector, we attract positive (yellow pixels) and repulse negative features (pink pixels).
% I think we can cut severely from this caption as the information is included in the main text and too low-level for a caption.
}
\label{fig:pixelsampling1}
\end{figure}

The training of the instance field in \emph{FruitNeRF++} is performed on a per-image basis, as the contrastive learning approach requires positive and negative pairs, which are only available within each image.
The task of the instance pixel sampler is to form the pixel pairs for training, as visualized in Fig. \ref{fig:pixelsampling1}.
To obtain positive pairs, we use the 2D instance mask to select multiple points from a single fruit, which are optimized to have similar instance embeddings.
By repeating the process above, we obtain groups of positive pairs. In turn, by matching pixels from different groups we obtain negative pairs which are optimized to be distinct through our contrastive loss (Eq. \ref{eq:contr}).
% In this case, both pixels get attracted in the feature space of the instance field.
% In contrast, a negative pixel pair is the combination with a pixel belonging to another fruit or the background. 
% Here we aim to repulse their feature vector.
Since it is harder to separate fruits that are (spatially) closer, we improve the training signal for such cases by mining local negatives (hard pairs).
To obtain these pairs, for each fruit we select $L-1$ nearest neighbours (in terms of distance between the 2D instance masks), and then sample $P$ pixels per fruit.
We generate $G$ such groups of fruits, which provide hard (local) negatives (Fig. \ref{fig:pixelsampling1}, right) and use all pairs formed from different groups as global negatives (Fig. \ref{fig:pixelsampling1} on the left).
% For the fruits we have on the one hand to guarantee that the feature space of the neighbouring fruits are fundamentally different (hard negatives) and on the other hand distant fruits also do not share the same feature space (weak negatives). 
% Therefore, we generate local and global negatives, as visualized in Fig. \ref{fig:pixelsampling} on the right. 
% For local negatives we select originating from our fruit $L$ nearest neighbour as local hard negatives and sample $P$ pixels per fruit.
% By generating $G$ different groups of fruits with pairs in their local vicinity, we automatically obtain corresponding global negatives (Fig. \ref{fig:pixelsampling} on the left).

\begin{table*}[t!] 
\definecolor{cellgreen}{RGB}{247,203,153}
%\fontsize{4pt}{4pt}\selectfont
\center
\caption{Fruit counting results on different fruit types with different models. The synthetic data have an image size of $1024~$px $\times$ $1024~$px and contain of 300 frames. 
Masks generated by \emph{Grounded-SAM}~\cite{ren2024grounded} have the corresponding fruit type as input. Only for plum we set the text prompt to `apple` \& `plum` and for mangoes to `apple`. 
As Detic~\cite{detic} classifies every object in the image, we select all available fruit classes. 
The results refer to counting result and are obtained if the center of a fruit is close to the GT center.
The best performing methods on each fruit is highlighted in green. $^{*}$: results are taken from \emph{FruitNeRF}~\cite{FruitNeRF}.} 
\begin{tabular}{l | c  c  c  c  | c  c  c  c   } 
\toprule
            & \multicolumn{4}{c|}{\emph{FruitNeRF}}                  & \multicolumn{4}{c}{\emph{FruitNeRF++}} \\
            &    \multicolumn{4}{c|}{\emph{GT Mask}}                 &  \multicolumn{4}{c}{\emph{GT Mask}} \\ \midrule  
Fruit Type  & {Count$^{*}$} & Precision$^{*}$ ($\uparrow$) & Recall$^{*}$ ($\uparrow$) &  F1-Score$^{*}$ ($\uparrow$)   & {Count}   & Precision ($\uparrow$) & Recall ($\uparrow$) & F1-Score ($\uparrow$)        \\ \midrule  
Apple       & 283/283       & 1         & 1      &  \cellcolor{cellgreen} 1       & 285/283   & 0.993     & 1      & \cellcolor{white} 0.996      \\             
Plum        & 642/745       & 0.973     & 0.812  & \cellcolor{cellgreen} 0.885    & 572/745   & 0.989     & 0.759  & \cellcolor{white} 0.859   \\          
Lemon       &  321/326      & 0.993     & 0.963  & \cellcolor{cellgreen} 0.978    & 312/326   & 1         & 0.957  &  \cellcolor{cellgreen} 0.978                      \\            
Pear        & 237/250       & 1         & 0.944  & \cellcolor{white} 0.971        & 252/250   & 0.992     & 1      &  \cellcolor{cellgreen} 0.996    \\           
Peach       & 148/152       & 1         & 0.973  & \cellcolor{white} 0.987        & 153/152   & 0.993     & 1      & \cellcolor{cellgreen} 0.996   \\       
Mango       &  929/1150     & 0.978     & 0.788  & \cellcolor{cellgreen} 0.873    & 649/1150  & 0.988     & 0.557  &  \cellcolor{white} 0.713                       \\\midrule        
            & \multicolumn{8}{c}{\emph{       FruitNeRF++}} \\
            &    \multicolumn{4}{c|}{\emph{SAM Mask}}                           &  \multicolumn{4}{c}{\emph{Detic Mask}}  \\ \midrule  
Fruit Type  & {Count}       & Precision ($\uparrow$) & Recall ($\uparrow$) &  F1-Score ($\uparrow$) & {Count}      & Precision ($\uparrow$) & Recall ($\uparrow$) & F1-Score ($\uparrow$)            \\ \midrule  
Apple       & 283/283       & 0.996     & 0.996  & 0.996                        &  284/283      & 0.996    & 1      & \cellcolor{cellgreen} 0.998                             \\             
Plum        & 254/745       & 1         & 0.341  & \cellcolor{cellgreen} 0.509                        &  137/745      & 1        & 0.184  & 0.311                          \\          
Lemon       & 307/326       & 0.997     & 0.938  & \cellcolor{cellgreen} 0.967                        &  274/326     & 0.993     & 0.834  & 0.907                                \\            
Pear        & 236/250       & 0.995     & 0.94   & \cellcolor{cellgreen} 0.967                        &  233/250     & 1         & 0.932  & 0.965                                \\           
Peach       & 152/152       & 0.993     & 0.993  & \cellcolor{cellgreen} 0.993                        &  150/152     & 0.987     & 0.974  & 0.98                               \\       
Mango       & 457/1150      & 0.987     & 0.392  & \cellcolor{cellgreen} 0.561                        &  324/1150    & 0.997     & 0.281  & 0.438                                \\\bottomrule        
\end{tabular} 
\label{tab:fruit_nerf_count}
\end{table*}

\if false
\begin{figure*}[t!] 
\definecolor{Worker1}{rgb}{0.26700401,  0.00487433,  0.32941519}
\definecolor{Worker2}{rgb}{0.2468114 ,  0.28323662,  0.53594093}
\definecolor{Worker3}{rgb}{0.14475863,  0.51909319,  0.55657181}
\definecolor{Worker4}{rgb}{0.18065314,  0.70140222,  0.48818938}
\definecolor{Worker5}{rgb}{0.55548397,  0.84025437,  0.26928147}
\definecolor{Worker6}{rgb}{0.99324789,  0.90615657,  0.1439362 }
     \centering 
     \begin{subfigure}{.32\linewidth} 
        \centering 
        \begin{tikzpicture}
	\begin{axis}[colormap/viridis,
                width=\linewidth,
                height=0.7\linewidth,
                grid=major,
                font=\tiny,
                %legend columns = 2,
                %legend style = {nodes={scale=0.8, transform shape}, at={(0.5, 1.25)}, anchor=north, inner sep=3pt, style={column sep=0.15cm}},       %vorher: at={(1.3, 1)}
                %legend cell align=left,
                %ticklabel style={
                %    /pgf/number format/.cd,
                %    fixed,
                %    use comma,
                %},
                xtick pos=bottom,
                ytick pos=right,
                xtick={1,2,3,4,5,6,7,8, 9, 10, 11},
                xticklabels={1, 2, 4, 6, 8, 16, 32, 64, 128, 192, 256},
                ymax=1.1,
                ytick={0, 0.2, 0.4, 0.6, 0.8, 1},
                ylabel={F1-Score},
                xlabel={Feature Embedding Size $D$},
	]
    \pgfplotsset{major grid style={dashed, color=black!20}} % modifies the style `every minor grid' 

    \draw[line width=0.01 mm] (-1000,280) -- (10000,280);
    \draw[line width=0.01 mm, style=dashed] (-1000,250) -- (10000,250);

    %     \addlegendimage{black, only marks, mark=square} % or mark=none?

    %\addlegendentry{Apple: $\lambda_c=1, \lambda_e=1 $}
    %\addlegendimage{black, only marks, mark=square} % or mark=none?
    %
    %\addlegendentry{Apple: $\lambda_c=1, \lambda_e=0 $}
    %\addlegendimage{black, only marks, mark=square} % or mark=none?

    %\addlegendimage{black, only marks, mark=square} % or mark=none?
    %\addlegendentry{\makebox[0pt][l]{Plum: $\lambda_c=1, \lambda_e=1 $}}

    %\addlegendimage{black, only marks, mark=square} % or mark=none?
    %\addlegendentry{\makebox[0pt][l]{Plum: $\lambda_c=1, \lambda_e=0 $}}

    \addplot[mark=square*, style=solid, color=Worker1, line width=0.15mm] coordinates { % Apple
		(1, 0.0132) % 3, 3 
		(2,0.1019) % 192, 202 
		(3,0.6928) % 198, 216
		(4,0.8627) % 247, 252
		(5,0.9094) % 275, 273
		(6,0.974) % 281, 282
		(7,0.9775) % 283, 283
		(8,0.9607) % 283, 283
  		(9,0.9675) % 283, 283
  		(10,0.9775) 
  		(11,0.9878)
	};
    \label{plot:apple_2}

    \addplot[mark=diamond*,  style=solid, color=Worker3, line width=0.15mm] coordinates {
		(1, 0.002614) % Plum
		(2, 0.1443) 
		(3, 0.4794) 
		(4, 0.623) 
		(5, 0.6948) 
		(6, 0.6602) 
		(7,0.7098) 
		(8,0.8387) 
  		(9,0.8748) 
  		(10,0.8865) 
  		(11,0.8705) 
	};
    \label{plot:plum_2}

    \addplot[mark=*,  style=solid, color=Worker5, line width=0.15mm] coordinates {
		(1, 0.001709) % Mango
		(2, 0.09156) 
		(3, 0.2915) 
		(4, 0.4392) 
		(5, 0.4413) 
		(6, 0.5402) 
		(7, 0.5907) 
		(8, 0.6247) 
  		(9,0.6081) 
  		(10,0.6274) 
  		(11,0.6553) 

 };
    \label{plot:mango_2}

\path[fill=red, fill opacity=0.5]  (55,200) -- (105,200) -- (105,-100) -- (55,-100) -- cycle;
\path[fill=red, fill opacity=0.2]  (45,200) -- (55,200)  -- (55,-100) -- (45,-100) -- cycle;

    %\coordinate (legend) at (axis description cs:0.88,1);

	\end{axis}     

    %\matrix [
    %    draw,
    %    matrix of nodes,
    %    anchor=south east,
    %    nodes={scale=0.8, transform shape}
    %] at (legend) {
    %   Apple: & \ref{plot:apple_1} $\lambda_c=1, \lambda_e=1 $ & \ref{plot:apple_2} $\lambda_c=1, \lambda_e=0 $ \\
    %   Plum:  & \ref{plot:plum_1} $\lambda_c=1, \lambda_e=1 $ & \ref{plot:plum_2} $\lambda_c=1, \lambda_e=0 $  \\
    %   Mango:  & \ref{plot:mango_1} $\lambda_c=1, \lambda_e=1 $ & \ref{plot:mango_2} $\lambda_c=1, \lambda_e=0 $  \\
    %};

\end{tikzpicture}
        \caption{Counting result on variation of the feature embedding size. Temperature is $\tau = 0.2$, $\lambda_e=0$ and $\lambda_c=1$. The red area describes the feasible area of feature embedding size $D$.}
        % As a basis we took a pre-trained density and semantic field and changed the output size of the instance field. Therefore we trained the network on the synthetic apple dataset for an additional 45000 iterations and evaluated the counting result for all runs. We evaluated the counted fruits for two different clustering metrics. One only on the cosine distance and the other one setting euclidean and cosine distance in a equilibrium.
        \label{subfig:feature_vector_dim}
     \end{subfigure}
     \begin{subfigure}{.32\linewidth} 
         \centering 
         \begin{tikzpicture}
\begin{axis}[
        width=\linewidth,
        height=0.7\linewidth,
        grid=major,
        font=\tiny,
        legend style={
            at={(0.95,0.1)}, anchor=north east, style={ column sep=.15cm}},
        xtick pos=bottom,
        ytick pos=right,
        xtick={1,2,3,4,5,6,7,8, 9},
        xticklabels={.0125, .025, .05, 0.1, 0.2, 0.3, 0.4, 0.5, 0.6},
        ymax=1.1,
        ymin=0,
        ytick={0, 0.2, 0.4, 0.6, 0.8, 1},
        ylabel={F1-Score},
        xlabel={Temperature $\tau$},
]
\pgfplotsset{major grid style={dashed, color=black!20}} % modifies the style `every minor grid' 

%\draw[line width=0.01 mm] (-1000,280) -- (10000,280);

%\draw[line width=0.01 mm, style=dashed] (-1000,250) -- (10000,250);

\path[fill=red, fill opacity=0.2]  (250,200) -- (350,200) -- (350,-100) -- (250,-100) -- cycle;
\path[fill=red, fill opacity=0.5]  (350,200) -- (550,200) -- (550,-100) -- (350,-100) -- cycle;
\path[fill=red, fill opacity=0.2]  (550,200) -- (650,200) -- (650,-100) -- (550,-100) -- cycle;

\addplot[mark=square*, color=Worker1, style=solid, line width=0.15mm] coordinates {
    (1,0.7255) % Apple
    (2,0.8219)
    (3,0.3631)
    (4,0.8998)
    (5,0.9639)
    (6,0.8799)
    (7,0.699)
    (8,0.5954)
    (9,0.479)
};
\label{plot:temperature_apple}

\addplot[mark=diamond*, color=Worker3, style=solid, line width=0.15mm, mark size=3pt] coordinates {
    (1,0.736) % Plum
    (2,0.6515)
    (3,0.1307)
    (4,0.3976)
    (5,0.7126)
    (6,0.6853)
    (7,0.6545)
    (8,0.6341)
    (9,0.5167)
};
\label{plot:temperature_plum}

\addplot[mark=*, color=Worker5, style=solid, line width=0.15mm] coordinates {
    (1,0.3965) % Mango
    (2,0.08488)
    (3,0.3593)
    (4,0.5108)
    (5,0.5979)
    (6,0.5295)
    (7,0.5144)
    (8,0.4868)
    (9,0.4399)
};
\label{plot:temperature_mango}

%\addlegendentry{Apple}
%\addlegendentry{Plum}
%\addlegendentry{Mango}

    \coordinate (legend) at (axis description cs:1,-0.6);

\end{axis}     
   \matrix [
       draw,
       matrix of nodes,
       anchor=south east,
       nodes={scale=0.6, transform shape}
   ] at (legend) {
      Apple: & \ref{plot:temperature_apple}, & Plum: \ref{plot:temperature_plum}, & Mango: \ref{plot:temperature_mango} \\
   };

\end{tikzpicture}
         \caption{Sweep over temperature $\tau$. Here we set $\lambda_c = 1$ and $\lambda_e = 0$}
         \vspace{0.28cm}
    \label{subfig:temperature}
     \end{subfigure} 
    \begin{subfigure}{.32\linewidth} 
         \centering 
         \begin{tikzpicture}
	\begin{axis}[
                    width=\linewidth,
                    height=0.7\linewidth,
                    grid=major,
                    font=\tiny,
                    legend style={
                        at={(0.95,0.1)}, anchor=south east, style={ column sep=.15cm}},
                    xtick pos=bottom,
                    ytick pos=right,
                    xmax=10.5,
                    xmin=-0.5,
                    ymax=1.2,
                    ylabel={F1-Score},
                    xlabel={$\lambda_e$},
                    %style={line width=0.25mm}
	]
    \pgfplotsset{major grid style={dashed, color=black!20}} % modifies the style `every minor grid' 

    %\draw[line width=0.01 mm] (-1000,280) -- (10000,280);

    %\draw[line width=0.01 mm, style=dashed] (-1000,250) -- (10000,250);

	\addplot[mark=square*, color=Worker1, line width=0.15mm] coordinates {
		(0,0.96385542) % Apple
		(1,0.98434783)
		(2,0.98951049)
		(3,0.98951049)
		(4,0.98606272)
		(5,0.98606272)
		(6,0.99647887)
		(7,0.99124343)
  		(8,0.99124343)
      	(9,0.99124343)
  		(10,0.98951049)
	};

 	\addplot[mark=diamond*, color=Worker3, line width=0.15mm] coordinates {
		(0,0.71571906) % Plum
		(1,0.71571906)
		(2,0.88856729)
		(3,0.83596431)
		(4,0.87226535)
		(5,0.87226535)
		(6,0.88713156)
		(7,0.89146165)
  		(8, 0.88937093)
      	(9,0.89241877)
  		(10,0.88856729)
	};

	\addplot[mark=*, color=Worker5, line width=0.15mm] coordinates {
		(0,0.57938144) % Mango
		(1,0.76777251)
		(2,0.85262117)
		(3,0.75483871)
		(4,0.79312142)
		(5,0.81169831)
		(6,0.82615307)
		(7,0.83559578)
  		(8,0.84042021)
      	(9,0.83983984)
  		(10,0.8516129)
	};

 	\addplot[mark=*, color=Worker6, style=dotted] coordinates {
		(0,0) % Fuji
		(0.2,0.0)
		(0.4,0.0)
		(0.6,0.0)
		(0.8,0.0)
		(1.0,0.0)
		(1.2,0.0)
		(1.4,0.0)
  		(1.6,0.0)
      	(1.8,0.0)
  		(2.0,0.0)
  		(2.2,0.0)
  		(2.4,0.0)
  		(2.6,0.0)
  		(2.8,0.0)
      	(3.0,0.0)
	};

 	%\addlegendentry{Apple}
   	%\addlegendentry{Plum}
 	%\addlegendentry{Mango}
 	%\addlegendentry{Fuji}

	\end{axis}
\end{tikzpicture}
         \vspace{0.63cm}
        \caption{Cluster performance via parameter sweep over $\lambda_e$ with $\lambda_c=1$. $\tau=0.2$ and $D=32$.}
        \label{subfig:cluster_sweep}
     \end{subfigure} 
     \caption{
     %\todo[inline]{@Lukas: Write caption and fix figure layout. @All: any idea where to place the legend?
     }
\vspace{-0.3cm}
\end{figure*} 

\fi

\subsubsection{Cascaded Training}
\label{ssec:cascaded}
The training of the semantic and instance fields is unstable if the representation of density in the neural field is poor. 
Therefore, we use a cascaded training method, by optimizing at first only the density and appearance fields. Then, the semantic field is added and trained jointly. 
Finally, we freeze the appearance and semantic fields and train only the instance field; the other parts of the network are frozen, as the instance pixel sampler may otherwise cause catastrophic forgetting.

\subsection{Point Cloud Export}

\emph{FruitNeRF++} gathers information about density, semantics, and instance segmentation within its different neural radiance fields. 
To export a point cloud after training, we effectively sample each field in the pipeline by uniformly sampling the space of our scene. 
This volumetric sampling is achieved by querying the network with every spatial point to extract relevant information.

The density field provides precise details about a point's geometric significance, the appearance field captures its color composition, the fruit field indicates the presence of a fruit, and the instance field reveals the implicit affiliation to a specific fruit id. 
Since the density field contains information about the physical properties of the tree, and the semantic field identifies whether a point belongs to a fruit, we can combine both point clouds to obtain a fruit point cloud.

The instance point cloud, derived from sampling the instance field, includes a $D$-dimensional feature vector for each point, encoding the fruit instance. 
By combining the fruit point cloud with the instance point cloud, we create a point cloud that represents only the fruits, along with their corresponding $D$-dimensional feature vectors.

%By clustering this multi-dimensional point cloud, we can perform clustering not only in Euclidean space but also in feature space, making the approach more robust for fruits in close proximity.

\subsection{Fruit Counting via Multi-Modal Clustering}
\label{sec:counting_instance_vector}

% Our revisited fruit counting approach significantly deviates from the clustering concept used in \emph{FruitNeRF}~\cite{FruitNeRF}. 
Our fruit-counting approach is radically different from the clustering concept used in \emph{FruitNeRF}~\cite{FruitNeRF}.
% There, we employed a cascaded clustering procedure to separate clusters of fruits from single fruits. 
There, a cascaded clustering procedure is employed by separating first conglomerates of fruits from single fruits.
% A cluster of fruits is defined as a point cloud containing multiple fruits. 
% By clustering the fruit clusters and applying a fruit-specific template matching we obtain the count. 
Then, the conglomerates of fruits are subdivided and counted using another clustering step, based on a fruit-specific template.
However, a drawback of this method is that both stages (DBSCAN \cite{dbscan} and agglomerative clustering \cite{scikitlearn}) require several hyper-parameters, which heavily influence the results, making the approach impractical for a multi-fruit counting approach.
In \emph{FruitNeRF++}, we replace this procedure with a simpler clustering approach that is divided into spatial partitioning and multi-modal clustering.

\subsubsection{Spatial Partitioning}
Spatial partitioning uses $k$-Means \cite{scikitlearn} to divide the existing Euclidean fruit point cloud into $S$ different partitions with similar point cloud size. 
The value of $S$ is arbitrarily selected and simply aims to reduce the number of points in the multi-feature clustering step, enhancing computational efficiency.

\subsubsection{Multi-Modal Clustering}
The multi-feature clustering is applied to all $S$ partitions independently.
For each $S$, we cluster the point cloud using HDBSCAN~\cite{hdbscan} with a custom distance metric composed of an Euclidean and a cosine distance.
That is, given two points $P_k = (\vect{i}_k, \vect{x}_k)$ and  $P_l = (\vect{i}_l, \vect{x}_l)$, our metric computes the distance as follows:
\begin{equation}
    d_{kl} =  \lambda_c d_c(\vect{i}_k, \vect{i}_l) + \lambda_e d_e(\vect{x}_k, \vect{x}_l)\;\;.
\end{equation}
Here, $\vect{i} \in \mathbb{R}^D$ represents the feature embedding, and $\vect{x}$ the 3D position. 
$d_c$ is defined as the cosine distance:
$
     d_c(\vect{a}, \vect{b}) =  1 - \frac{\langle \vect{a}, \vect{b}\rangle}{\lVert \vect{a} \rVert  \lVert \vect{b} \rVert}
$
which measures the distance between two feature vectors and is weighted by $\lambda_c\!=\!1$. 
The Euclidean distance $d_e$ between two points is weighted by  $\lambda_e\!=\!5$. A detailed evaluation of feasible $\lambda_e$ is provided in Sec. \ref{sec:evaluation}.

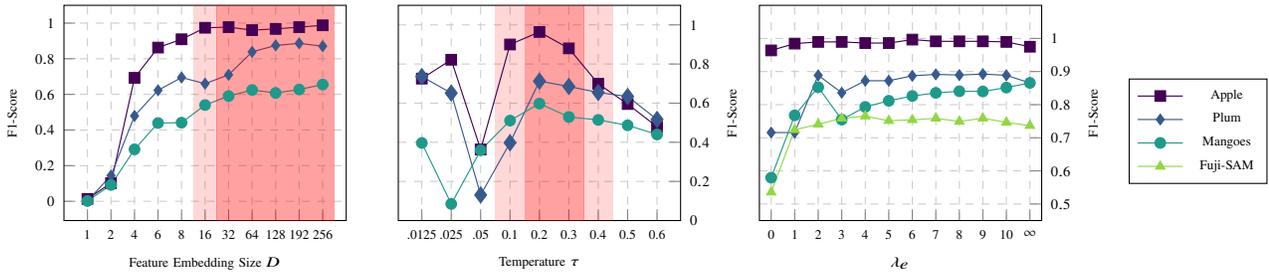
\begin{figure*}[t!]
   %\resizebox{\linewidth}{!}{%
   \center
        \begin{tikzpicture}
        \hspace{-0.28\linewidth}
	\begin{axis}[colormap/viridis,
                width=0.3\linewidth,
                height=0.25\linewidth,
                grid=major,
                font=\tiny,
                %legend columns = 2,
                %legend style = {nodes={scale=0.8, transform shape}, at={(0.5, 1.25)}, anchor=north, inner sep=3pt, style={column sep=0.15cm}},       %vorher: at={(1.3, 1)}
                %legend cell align=left,
                %ticklabel style={
                %    /pgf/number format/.cd,
                %    fixed,
                %    use comma,
                %},
                ytick pos=left,
                xtick pos=bottom,
                xtick={1,2,3,4,5,6,7,8, 9, 10, 11},
                xticklabels={1, 2, 4, 6, 8, 16, 32, 64, 128, 192, 256},
                ymax=1.1,
                ytick={0, 0.2, 0.4, 0.6, 0.8, 1},
                ylabel={F1-Score},
                xlabel={Feature Embedding Size $D$},
	]
    \pgfplotsset{major grid style={dashed, color=black!20}} % modifies the style `every minor grid' 

    \draw[line width=0.01 mm] (-1000,280) -- (10000,280);
    \draw[line width=0.01 mm, style=dashed] (-1000,250) -- (10000,250);

    \addplot[mark=square*, style=solid, color=Worker1, line width=0.15mm] coordinates { % Apple
		(1, 0.0132) (2,0.1019) (3,0.6928) (4,0.8627) (5,0.9094) (6,0.974)  (7,0.9775) (8,0.9607) (9,0.9675) (10,0.9775) 
  		(11,0.9878)
	};
    \label{plot:apple_2}

    \addplot[mark=diamond*,  style=solid, color=Worker3, line width=0.15mm] coordinates {
		(1, 0.002614) (2, 0.1443) (3, 0.4794) (4, 0.623) (5, 0.6948) (6, 0.6602) (7,0.7098) (8,0.8387) (9,0.8748) (10,0.8865) (11,0.8705) 
	};
    \label{plot:plum_2}

    \addplot[mark=*,  style=solid, color=Worker5, line width=0.15mm] coordinates {
		(1, 0.001709) (2, 0.09156)  (3, 0.2915)  (4, 0.4392)  (5, 0.4413)  (6, 0.5402)  (7, 0.5907)  (8, 0.6247)  (9,0.6081)  (10,0.6274)  (11,0.6553) 

 };
    \label{plot:mango_2}

\path[fill=red, fill opacity=0.37]  (55,200) -- (105,200) -- (105,-100) -- (55,-100) -- cycle;
\path[fill=red, fill opacity=0.16]  (45,200) -- (55,200)  -- (55,-100) -- (45,-100) -- cycle;

    %\coordinate (legend) at (axis description cs:0.88,1);

	\end{axis}    
    \hspace{0.25\linewidth} 
    \begin{axis}[
            width=0.3\linewidth,
            height=0.25\linewidth,
            grid=major,
            font=\tiny,
            legend style={
                at={(0.95,0.1)}, anchor=north east, style={ column sep=.15cm}},
            xtick pos=bottom,
            ytick pos=right,
            xtick={1,2,3,4,5,6,7,8, 9},
            xticklabels={.0125, .025, .05, 0.1, 0.2, 0.3, 0.4, 0.5, 0.6},
            ymax=1.1,
            ymin=0,
            ytick={0, 0.2, 0.4, 0.6, 0.8, 1},
            ylabel={F1-Score},
            xlabel={Temperature $\tau$},
    ]
    \pgfplotsset{major grid style={dashed, color=black!20}} % modifies the style `every minor grid' 
    
    %\draw[line width=0.01 mm] (-1000,280) -- (10000,280);
    
    %\draw[line width=0.01 mm, style=dashed] (-1000,250) -- (10000,250);
    
    \path[fill=red, fill opacity=0.16]  (250,200) -- (350,200) -- (350,-100) -- (250,-100) -- cycle;
    \path[fill=red, fill opacity=0.37]  (350,200) -- (550,200) -- (550,-100) -- (350,-100) -- cycle;
    \path[fill=red, fill opacity=0.16]  (550,200) -- (650,200) -- (650,-100) -- (550,-100) -- cycle;

    \addplot[mark=square*, color=Worker1, style=solid, line width=0.15mm] coordinates {
        (1,0.7255) (2,0.8219) (3,0.3631) (4,0.8998) (5,0.9639) (6,0.8799) (7,0.699) (8,0.5954) (9,0.479)
    };
    \label{plot:temperature_apple}
    
    \addplot[mark=diamond*, color=Worker3, style=solid, line width=0.15mm, mark size=3pt] coordinates {
        (1,0.736) (2,0.6515) (3,0.1307) (4,0.3976) (5,0.7126) (6,0.6853) (7,0.6545) (8,0.6341) (9,0.5167)
    };
    \label{plot:temperature_plum}

    \addplot[mark=*, color=Worker5, style=solid, line width=0.15mm] coordinates {
        (1,0.3965) (2,0.08488)(3,0.3593)(4,0.5108)(5,0.5979)(6,0.5295)(7,0.5144)(8,0.4868)(9,0.4399)
    };
    \label{plot:temperature_mango}

    %\addlegendentry{Apple}
    %\addlegendentry{Plum}
    %\addlegendentry{Mango}

    \end{axis}     
    \hspace{0.27\linewidth} 
	\begin{axis}[
                    width=0.3\linewidth,
                    height=0.25\linewidth,
                    grid=major,
                    font=\tiny,
                    legend style={
                        at={(1.8,0.17)}, anchor=south east, style={ column sep=.15cm}},
                    xtick pos=bottom,
                    ytick pos=right,
                    xmax=11.5,
                    xmin=-0.5,
                    xtick={0,1,...,11},
                    xticklabels={0,1,2,3,4,5,6,7,8,9,10,$\infty$},
                    ytick={0.5, 0.6,  0.7, 0.8, 0.9, 1},
                    ymax=1.1,
                    ymin=0.45,
                    ylabel={F1-Score},
                    xlabel={$\lambda_e$},
                    %style={line width=0.25mm}
	]
    \pgfplotsset{major grid style={dashed, color=black!20}} % modifies the style `every minor grid' 

	\addplot[mark=square*, color=Worker1, line width=0.15mm] coordinates {
		(0,0.96385542) (1,0.98434783) (2,0.98951049) (3,0.98951049) (4,0.98606272) (5,0.98606272) (6,0.99647887) (7,0.99124343) (8,0.99124343) (9,0.99124343) (10,0.98951049) (11,0.97491039)
	};
     \addlegendentry{Apple}

 	\addplot[mark=diamond*, color=Worker3, line width=0.15mm] coordinates {
		(0,0.71571906) (1,0.71571906)(2,0.88856729)(3,0.83596431)(4,0.87226535)(5,0.87226535)(6,0.88713156)(7,0.89146165)(8, 0.88937093)(9,0.89241877)(10,0.88856729) (11,0.86568627 )
	};
     \addlegendentry{Plum}

	\addplot[mark=*, color=Worker5, line width=0.15mm] coordinates {
		(0,0.57938144) (1,0.76777251) (2,0.85262117) (3,0.75483871) (4,0.79312142) (5,0.81169831) (6,0.82615307) (7,0.83559578) (8,0.84042021) (9,0.83983984) (10,0.8516129) (11,0.86568627 )
	};
      \addlegendentry{Mangoes}

 	\addplot[mark=triangle*, color=Worker6, line width=0.15mm] coordinates {
		(0,0.5355932203389833) (1, 0.724014336917563) (2, 0.7408460718094562) (3, 0.7587939698492464) (4,0.7651542649727769) (5, 0.7515460167333577 ) (6,0.7541103397880893) (7,0.7588495575221239) (8,0.7492647058823534) (9,0.7590048273301154)  (10,0.7464788732394368) (11, 0.73656716)
  };
     \addlegendentry{Fuji-SAM}

     %\coordinate (legend) at (axis description cs:0.9,0.22);

	\end{axis}
   %\matrix [
   %    draw,
   %    matrix of nodes,
   %    anchor=south east,
   %    nodes={scale=0.6, transform shape}
   %] at (legend) {
   %   Apple & \ref{plot:temperature_apple}, & Plum \ref{plot:temperature_plum}, & Mango \ref{plot:temperature_mango} & FUJI \ref{plot:temperature_mango} \\
   %};
    %\hspace{0.3\linewidth} 

   %\begin{tikzpicture}
   % \begin{axis}[
   %     hide axis,
   %                 width=0.0\linewidth,
   %                 height=0.1\linewidth,
   %     scale only axis,
   %     legend style={at={(1,1)},anchor=north west},
   %     %legend cell align={left},
   %     no markers,
   % ]
   % \addplot[color=red, mark=o] coordinates {(0,0)};
   % \addlegendentry{Red Line}
%
   % \addplot[color=blue, mark=x] coordinates {(0,0)};
   % \addlegendentry{Blue Line}
%
   % \end{axis}

    %\end{tikzpicture}

\end{tikzpicture}
        
        \caption{Counting result on variation of the embedding size $D$ is depicted on the left. 
        Temperature is set to $\tau = 0.2$. $\lambda_e=0$ and $\lambda_c=1$ to show the impact of the embedding size only. 
        In the middle the sweep over temperature $\tau$ is shown.
        Here we set $\lambda_e = 1$ and $\lambda_c = 1$.
        On the right shows the cluster performance via parameter sweep over $\lambda_e$ while setting $\lambda_c=1$.
        The red area describes the feasible area for parameters $D$ and $\tau$. 
}
        \label{fig:evaluation}
        \vspace{-0.8em}
\end{figure*}

\if false
\section{Implementation Details}

\subsection{Instance Segmentation}
\label{sub:instance_segmentation_implementation}
\todo[inline]{@Lukas: Merge into Fruit Segmentation section}

To achieve a unified fruit counting approach we utilize \emph{Grounded-SAM} \cite{ren2024grounded} and Detic \cite{detic} to predict precise instance segmentation masks without training a custom model.

For \emph{Grounded-SAM} \cite{ren2024grounded}, we use the default model structure and pre-trained weights of \emph{Grounding DINO} \cite{liu2023grounding}.
For SAM \cite{kirillov2023segany} we use the weights of SAM-HQ \cite{samhq}.  
For obtaining the instance masks, we use the name of each fruit as a text prompt, except for the fruit mango, where we added `apple' as a text prompt due to poor masks with `mango' only.
For large images like the FUJI dataset \cite{fujiapple}, we split each image into four equally sized overlapping images to obtain more bounding boxes provided by \emph{Grounding DINO}.
On average the SAM generated masks have an intersection over Union (IoU) of $0.561$.

For Detic \cite{detic} we choose the default network configuration and selected LVIS (Large Vocabulary Instance
Segmentation) \cite{lvis} as a vocabulary collection.
As Detic detects every class in an image given by the vocabulary collection, the predictions are filtered in a post-processing step by all fruit-related classes available in LVIS.
On Tab. \ref{tab:segmentationresult} the full list of vocabulary classes are listed.
The predicted masks of Detic have an average IoU of $0.49$ over all fruits. 
 
\fi

\if false
\begin{table}[t!]
    \centering
    \caption{For the Detic we took an array of multiple fruit class present in the Lvis Vocabulary and applied it to all fruit images. Listet tags:  apple, apricot orange\_(fruit), peach, persimmon, mandarin\_orange, pear, banana, mango, lemon, pumpkin, plum, grape, cherry, blackberry, fig\_(fruit), blueberry, pinecone, raspberry, date\_(fruit), almond, lime, clementine. }
    \label{tab:segmentationresult}

    \begin{tabular}{l | c c | c c}
       IoU &  SAM      &  Prompt        & Detic   & Classes   \\ \toprule
     apple &  0.659    &  apple         & 0.632   & \multirowcell{6}{see\\ caption}  \\
     pear  &  0.519    &  pear          & 0.468   &    \\
     plum  &  0.382    &  plum          & 0.144   &    \\
     lemon &  0.556    &  lemon         & 0.514   &    \\
     peach &  0.662    &  peach         & 0.654   &    \\
     mango &  0.588    &  mango + apple & 0.528   &    \\
    
    \end{tabular}

\end{table}
\fi

\if false

% Our experiments utilize both synthetic data for various fruits and real-world data for benchmarking.
We evaluate our method on both synthetic and real-world data.
The synthetic dataset incorporates the models from \emph{FruitNeRF}, consisting of different fruit trees \cite{fruitassets}: apple, plum, lemon, pear, peach, and mango. 
For each tree model, we render 300 images by randomly placing a virtual camera looking towards the tree center, and extract the extrinsic and intrinsic camera parameters using the BlenderNeRF plugin \cite{BlenderNeRF}.

In addition to rendering color and semantic images, we extend the dataset by also rendering the ground-truth instance masks (GT Mask) of the fruits for each tree model. 
To enable the detection of fruits, we render the images at high resolution ($2048$\,px $\times$ $2048$\,px); however, a downsampled size ($1024$\,px $\times 1024$\,px) is sufficient for training the NeRF.
% To increase the segmentation quality for Grounded-SAM and Detic, we rendered the images at a resolution of $2048$\,px $\times$ $2048$\,px with a focal length of $35$\,mm and applied each model to this data. 
% For training the NeRF, we then used a downsampled version of $1024$\,px $\times 1024$\,px. 
A visualization of the dataset is shown on the project website.

The FUJI \cite{fujiapple} dataset consists of $582$ images taken from 12~trees (291~images per side) at a commercial apple orchard. 
The captured images have a resolution of $5184$\,px $\times$\, $3456$\,px and a focal length of $24$\,mm.

To recover the camera poses, we applied COLMAP \cite{schoenberger2016sfm} to each side, manually registered both sides, and further refined the result using ICP \cite{ICP}. The semantic and instance masks were generated using the visual foundation models Grounded-SAM and Detic.

All images for training were downsampled to a resolution of $1296$\,px $\times$ $864$\,px.

\fi

% \subsection{Nearest Fruit Graph / K Nearest Fruits}
% \label{sec:nearest_fruit_graph}

% Before using the instance masks for training the positional relationships between the fruits on a per image level have to be stored.
% This is archived by firstly computing the median pixel location of all available fruits.
% Afterwards, the euclidean distance between all fruit centers are computed.
% By sorting it by the nearest neighbour we are able to computed local fruit clusters which is relevant for sampling instance pixels according to their local and global neighbours. Image pixels are discussed in the following section.

\if false

\subsection{FruitNeRF++ Implementation and Training}
\label{sec:implementationandtraining}
We implemented \emph{FruitNeRF++} within the open-source neural radiance framework \emph{Nerfstudio} \cite{nerfstudio}.
The model foundation is provided by Nerfacto, which is a combination of different techniques to enhance the reconstruction quality on real data. 
In this section, we discuss the elements of \emph{FruitNeRF++} and refer to our model architecture, the utilized loss function and training procedure. 

\subsubsection{Network Architecture}
\label{ssec:architecture}
Our neural field structure is based on InstantNGP \cite{instantngp}. 
% The base-MLP for density takes a 3D coordinate $\mathbf{x}$ as an input to a two-layered MLP with a hidden dimension of 128 neurons that outputs a geometric feature-embedding $\mathbf{\sigma}$ of dimension 30. 
The 3D coordinate is processed by a hash encoding followed by a two-layer MLP which outputs a geometric feature embedding.
% For the color  we concatenate the viewing direction $\mathbf{d}$ with the geometric feature embedding and query it through a three-layered MLP with a width of 128 neurons.
The color $\mathbf{\hat{C}}$ is computed by a three-layer MLP which takes as input the geometric features and the view direction.
% Both $\mathbf{x}$ and $\mathbf{d}$ are encoded using frequency encoding \cite{nerf}.
% The semantic field takes only the geometric embedding $\mathbf{\sigma}$ as an input, and a MLP with a depth of two and a width of 128 predicts a semantic scalar $s$.
The semantic branch is a two-layer MLP that takes the geometric embedding as input and outputs a single scalar: the fruit likelihood.
% For the instance field we use a 5-layer MLP with a width of 128 neurons that predicts a $D$-dimensional instance feature embedding $\mathbf{i}$ giving the geometric embedding and a spatial coordinate $\mathbf{x}$.
% Similarly to Bhalgat et al. \cite{contrastivelift}, we only use the raw 3D coordinate and no frequency encoding, as most spatial information is already encoded in the feature embedding.
The instance field uses a 5-layer MLP to predict $D$-dimensional instance embeddings $\mathbf{i}$ given the geometric embedding an the spatial coordinate $\mathbf{x}$.
% Probably can be removed vvv
% The instance field output uses \textit{Tanh} as an output activation function to squash the embeddings between $$-1$ and $1$.

\fi

% \todo[inline]{@Lukas: add pixel sampler details}

%\subsection{Volumetric Field Sampling}

\if false
\subsection{Instance Visualization \& Viewer}
\label{sec:instance_visualizer_and_viewer}

\todo[inline]{@Lukas: This doesn't seem critical for the understanding of the method; maybe we can save some space.}

For visualization of the instance masks we employed two different methods for training and inference time. 
During training we can not readily (without further ado) evaluate the instance masks as we do not want to disrupt the training process by clustering the feature point cloud repeatedly.
Therefor it is sufficient to visualize if a feature vectors on a pixel has a similar value as points in its near vicinity. 
Thus we generate a random projection matrix $P\in \mathbb{R}^{N\times 3}$ to map the $N$-dimensional feature vector onto a RGB color value.
In the Viewer each feature space gets mapped to a RGB color space.
Due to the random mapping the color sometimes appear ambiguous.

For rendering of predicted instance masks inside the viewer we extract all median cluster centers of the instance feature vectors point cloud (after clustering). 
For every rendering of an image we assign each projected feature vectors to the nearest cluster center and assign them a unique color value.
To determine the nearest cluster we use only the cosine distance. 
Thus we now can visualize multi-view consistent instance masks.
\fi

\section{Evaluation and Results}
\label{sec:evaluation}

%\begin{itemize}
 %   \item F1-Score of count of synthetic fruit trees 
 %   \item F1-Score of count of Fuji-Dataset \cite{fujiapple}
 %   \item Estimate fruit count on both DBSCAN vs HDBSCAN
 %   \item ROC Analysis on Fuji-Dataset. Varying Network size/Clustering parameters/feature vector size in relation to fruit count?
 %   \item Size estimation and comparison to GT (both synthetic and real)
 %   \item Panoptic Quality (PQ) - Metric \cite{POSeg}?
 %   \item Compare to SoTA (Panoptic Lifting, Contrastive Lift)
%\end{itemize}

%\subsubsection{Online Instance Count Metric - OICM}

%\begin{itemize}
%    \item Linear probe suggestion vom Tomotai used during training of instance field
%    \item Linear layer with num feature size es input and output size equals num classes (without background as bg has to many different features (no direct optimization on background points)). Afterward cross entropy loss and predict class/instance id
%    \item At evaluation steps a the F1/precision/recall score on a per pixel level is evaluated (only on pixels which belong according to synthetic data to a fruit) --> maybe take union of predicted and target semantics for computation to look at false positive?
%    \item At evaluation steps a the F1/precision/recall score on the fruit count is evaluated. Compares predicted labels in an entire image with the target/gt instance id in the same image. 
%\end{itemize}

\if false
\subsection{Dataset}
\label{ssec:dataset}

Our experiments utilize both synthetic data for various fruits and real-world data for benchmarking.
The synthetic dataset incorporates the models from \emph{FruitNeRF}, consisting of different fruit trees \cite{fruitassets}: apple, plum, lemon, pear, peach, and mango. 
For each tree model, we render 300 images by randomly placing a virtual camera looking towards the tree center, and extract the extrinsic and intrinsic camera parameters using the BlenderNeRF plugin \cite{BlenderNeRF}.

In addition to rendering color and semantic images, we extend the dataset by also rendering the ground-truth instance masks (GT Mask) of the fruits for each tree model. 
To enable the detection of fruits, we render the images at high resolution ($2048$\,px $\times$ $2048$\,px); however, a downsampled size ($1024$\,px $\times 1024$\,px) is sufficient for training the NeRF.
% To increase the segmentation quality for Grounded-SAM and Detic, we rendered the images at a resolution of $2048$\,px $\times$ $2048$\,px with a focal length of $35$\,mm and applied each model to this data. 
% For training the NeRF, we then used a downsampled version of $1024$\,px $\times 1024$\,px. 
A visualization of the dataset is shown on the project website.

The FUJI \cite{fujiapple} dataset consists of $582$ images taken from 12~trees (291~images per side) at a commercial apple orchard. 
The captured images have a resolution of $5184$\,px $\times$\, $3456$\,px and a focal length of $24$\,mm.

To recover the camera poses, we applied COLMAP \cite{schoenberger2016sfm} to each side, manually registered both sides, and further refined the result using ICP \cite{ICP}. The semantic and instance masks were generated using the visual foundation models Grounded-SAM and Detic.

All images for training were downsampled to a resolution of $1296$\,px $\times$ $864$\,px.

\fi 

%\subsection{FruitNeRF++ Evaluation and Results}
%\label{sec:fruitcounting}

%The evaluation on \emph{FruitNeRF++} is done one the different synthetic data the performance of the neural radiance field itself by sweeping over the output instance embedding size $D$, the temperature $\tau$ from the contrastive loss function in Eq. \ref{eq:contr} and the euclidean weighting $\lambda_e$ for the clustering approach.
%Additionally we evaluate the clustering approach by sweeping over the euclidean weighting $\lambda_e$.
%We aim top find a feasible set of parameters which work on the selected datasets apples, plums and mangoes due to their varying number of fruits.

We evaluate \emph{FruitNeRF++} on our synthetic dataset. We used the same parameters for all experiments with an instance embedding dimension of $D\!=\!32$, a temperature of $\tau\!=\!0.2$ and $\lambda_c\!=\!1$ and $\lambda_e\!=\!1$.
In Tab. \ref{tab:fruit_nerf_count} the results show that our approach outperforms \emph{FruitNeRF} on most fruit types. 
Overall, \emph{FruitNeRF++} with ground truth masks achieve an average F1-score of 0.925 compared to \emph{Grounded-SAM}-generated masks at 0.832 and Detic masks at 0.776.

Secondly, we evaluate the impact of different feature embedding sizes $D$, which define the space that encodes the fruit identity.
In this experiment we sweep from $D\!=\!1$ to $D\!=\!256$, set the temperature to $\tau\!=\!0.2$ and the weighting parameters to $\lambda_e\!=\!0$ and $\lambda_c\!=\!1$.
In Fig. \ref{fig:evaluation} on the left the results for all three fruits are visualized. 
It can be seen that increasing the feature dimension leads to an increase in F1-Score as more fruit identities can be encoded.
For apples, the F1-Score converges to $1$ at $D\!=\!16$, whereas for plums and mangoes the maximum is reached with a larger feature embedding size.
The poor result for mangoes can be attributed to the large amount of fruits and high occlusions inside the tree.

Our next experiment looks at the contrastive loss function and its temperature parameter $\tau$. 
% , which controls the strength of penalties on hard negative samples. 
A lower temperature increases the penalty on the hardest negatives, leading to more separation between local structures \cite{cl_behaviour} and causes the loss to concentrate on points in its vicinity.
In contrast, a higher (e.g. $\tau\! \rightarrow\! +\infty$) temperature reduces sensitivity to hard negatives and gives all negative points the same magnitude of penalties.
A feasible $\tau$ aims for a reasonable distribution which is locally clustered and globally separated.
From Fig. \ref{fig:evaluation}, center, it is evident that the feasible region is between 0.1 and 0.3 and peaks at $0.2$. 
In the edge cases, the clustering is not able to properly separate the instance embeddings.

To investigate our custom clustering metric (Fig. \ref{fig:evaluation}, right), we fixed $\lambda_c\!=1\!$ and varied $\lambda_e$ from 0 to 10. 
The case $\lambda_e\!=\!\infty$ indicates $\lambda_c\!=\!0$ and $\lambda_e\!=\!1$. 
The clustering performance with only cosine distance is limited, yet increasing rapidly when combined with the Euclidean component. Removing the cosine component similarly results in a drop in performance.
% Without an Euclidean component the clustering approach is able to separate a certain amount of fruits.
% With increasing $\lambda_e$ the clustering results further improves.
The minor contribution of feature embeddings to the clustering results (e.g. for apple) can be can be attributed to the fact that for synthetic data clustered fruits occur rarely and thus the main separation can be done in Euclidean space. 

For the FUJI dataset, our method \emph{FruitNeRF++} obtains an F1-Score of $0.765$ using instance masks generated by \emph{Grounded-SAM}, an embedding size of $D=32$, temperature $\tau\!=\!0.35$ and $\lambda_e\!=\!5$. We sampled 600k points and chose $S\!=\!40$ partitions.
%For the original instance masks we obtained an F1-Score of \textcolor{red}{$?$} with the same parameters. 
The results of our generalized multi-fruit approach compare well to the fruit-specific approach from Gené-Mola et al.~\cite{fuji_count} with an F1-Score of $0.881$.
The results of our counting algorithm compared to Gené-Mola et al.~\cite{fuji_count} can be attributed to the noisy poses and manual registration of both sides.

% https://tex.stackexchange.com/questions/43832/how-can-i-create-bar-plot-groups-of-different-sizes-in-pgfplots

%\begin{figure}[t]
%    \centering
%    \includegraphics[width=1\linewidth]{figures/myplot.png}
%    \caption{FUJI. $\lambda_c = 1$. Title is y-label}
%    \label{fig:avc}
%\end{figure}

%\begin{figure}[h]
%\centering
%    \includegraphics[width=0.99\linewidth]{figures/cluster_vis/t-SNE_apple.pdf}
%    \caption{Visualization of the apple clusters of the synthetic apple tree. The feature vectors are projected to two dimensions using t-SNE}
%\end{figure}

\section{Limitations}

Our fruit counting method, \emph{FruitNeRF++}, has limitations. 
Training on larger scenes, like the Fuji dataset, takes about 8 hours on an Nvidia A5000 (24GB VRAM), making it unsuitable for real-time use.
The slow convergence of the instance field accounts for more than half of the training time.
Additionally the results heavily depend on a good coverage of the tree, and noisy data significantly decreases the correctness of the implicit field and thus the counting result.
This is mainly due to inaccurate poses, which cause the embeddings of one fruit to contaminate the neighbouring features.
Lastly, the instance segmentation tends to predict two different instances if a leaf or branch visually separates a fruit; this causes the feature embeddings of the two halves to be pulled apart.

As we claim to provide a generalized fruit counting approach, challenges arise for fruits that grow in clusters, such as bananas, berries, or dates.
Further research is needed to efficiently estimate cluster sizes and predict accurate cluster counts.

\section{Conclusion \& Outlook}
%\todo[inline]{@Lukas: evaluate messy (100 objects) room dataset scene!}

We introduced \emph{FruitNeRF++}, a novel method to count fruits from view-inconsistent instance segmentation masks using neural radiance fields. 
It encodes the instance identity of each fruit implicitly in the neural field and allows a simple clustering approach to obtain a fruit count. 
Compared to its predecessor \emph{FruitNeRF}, our approach is fully agnostic to the type of fruit, regarding both the neural radiance field and the clustering. 
This also applies to any type of object as it does not need any object shape priors.
In an additional experiment with the Messy Rooms dataset dataset \cite{contrastivelift} that contains 100 common household objects, we detected 99 objects. For a visualization visit our project website.

Recent work by Xu \emph{et al.}~\cite{Xu_2023_CVPR} introduces a class-agnostic object counting approach that requires only a class name to estimate object counts automatically. However, as this method is currently designed for single 2D images, extending it to 3D or multi-view images could be a promising direction, potentially improving accuracy and robustness in fruit counting.

As a future direction for a NeRF-based counting approach, it is necessary to reduce the computational complexity. Using time-series images and incrementally building up the scene with SLAM \cite{blenderslam}, Gaussian Splatting \cite{gssplatting} or PAgNeRF \cite{pagnerf} could help achieve real-time performance. Additionally, temporal information could simplify fruit ID association and further reduce computation time.

\section*{Acknowledgement}
We extend our gratitude to \textbf{Adam Kalisz} for his unique Blender skills, \textbf{Jann-Ole Henningson} for proof reading our script, \textbf{Victoria Schmidt}, and \textbf{Annika Killer} for their invaluable assistance in evaluating the recorded apple trees.

%%%%%%%%%%%%%%%%%%%%%%%%%%%%% Bibliography  %%%%%%%%%%%%%%%%%%%%%%%%%%%%


\begin{thebibliography}{99}

% Introduction
\bibitem{fcsota}
J. C. Miranda, J. Gené-Mola, M. Zude-Sasse, N. Tsoulias, A. Escolà, J. Arnó, J. R. Rosell-Polo, R. Sanz-Cortiella, J. A. Martínez-Casasnovas and E. Gregorio, "Fruit sizing using AI: A review of methods and challenges," \emph{Postharvest Biology and Technology}, 2023.


\bibitem{fuji_count}
J. Gené-Mola, R. Sanz-Cortiella, J. R. Rosell-Polo, J. R. Morros, J. Ruiz-Hidalgo, V. Vilaplana and E. Gregorio, "Fruit detection and 3D location using instance segmentation neural networks and structure-from-motion photogrammetry," \emph{Computers and Electronics in Agriculture}, 2020.

\bibitem{robust_fruit_counting}
X. Liu, S. W. Chen, S. Aditya, N. Sivakumar, S. Dcunha, C. Qu, C. J. Taylor, J. Das and V. Kumar, "Robust Fruit Counting: Combining Deep Learning, Tracking, and Structure from Motion," \emph{arXiv}, 2018.

\bibitem{liu2019monocular}
X. Liu et al., "Monocular Camera Based Fruit Counting and Mapping With Semantic Data Association," \emph{IEEE Robotics and Automation Letters}, vol. 4, pp. 2296-2303, 2019.

\bibitem{hanI2020comparative}
N. Häni, P. Roy and V. Isler, "A comparative study of fruit detection and counting methods for yield mapping in apple orchards," \emph{Journal of Field Robotics}, vol. 37, pp. 263–282, 2019.

\bibitem{FruitNeRF}
L. Meyer, A. Gilson, U. Schmidt and M. Stamminger, "\emph{FruitNeRF}: A Unified Neural Radiance Field based Fruit Counting Framework," \emph{arXiv}, 2024.

% Relates Work - Fruit Counting

\bibitem{pagnerf}
S. Claus, M. Halstead, P. Zimmer, T. Laebe, E. Guclu, C. Stachniss, and C. McCool, "PAg-NeRF: Towards fast and efficient end-to-end panoptic 3D representations for agricultural robotics," \emph{IEEE Robotics and Automation Letters}, 2023.

\bibitem{hqstrawberry}
A. Riccardi, S. Kelly, E. Marks, F. Magistri, T. Guadagnino, J. Behley, M. Bennewitz, and C. Stachniss, "Fruit Tracking Over Time Using High-Precision Point Clouds," \emph{Proceedings of the IEEE International Conference on Robotics and Automation (ICRA)}, 2023.

\bibitem{berryharvest}
M. Sorour, P. J. From, K. Elgeneidy, S. Kanarachos, and M. Sallam, "Compact Strawberry Harvesting Tube Employing Laser Cutter," in \emph{Proceedings of the IEEE/RSJ International Conference on Intelligent Robots and Systems (IROS)}, 2022.

\bibitem{unet}
O. Ronneberger, P. Fischer and T. Brox, "U-Net: Convolutional Networks for Biomedical Image Segmentation," \emph{Medical Image Computing and Computer-Assisted Intervention – MICCAI 2015}, N. Navab, J. Hornegger, vol. 9351, pp. 234–241, 2015.

\bibitem{ZHANG2022107062}
Y. Zhang, W. Zhang, J. Yu, L. He, J. Chen, and Y. He, "Complete and accurate holly fruit counting using YOLOX object detection,"  \emph{Computers and Electronics in Agriculture}, vol. 198, p. 107062, 2022. 

% Relates Work - Contrastive learniong 
\bibitem{hadsell2006dimensionality}
R. Hadsell, S. Chopra, and Y. LeCun, "Dimensionality Reduction by Learning an Invariant Mapping," \emph{2006 IEEE Computer Society Conference on Computer Vision and Pattern Recognition (CVPR)}, 2006.

\bibitem{simclr}
T. Chen, S. Kornblith, M. Norouzi, and G. Hinton, "A Simple Framework for Contrastive Learning of Visual Representations," \emph{Proceedings of the 37th International Conference on Machine Learning (ICML)}, 2020.

\bibitem{stego}
M. Hamilton, Z. Zhang, B. Hariharan, N. Snavely, and W. T. Freeman, "Unsupervised Semantic Segmentation by Distilling Feature Correspondences," \emph{International Conference on Learning Representations}, 2022.

\bibitem{cadet}
C. Guille-Escuret, P. Rodriguez, D. Vazquez, I. Mitliagkas, and J. Monteiro, "CADet: Fully Self-Supervised Out-Of-Distribution Detection With Contrastive Learning," \emph{Advances in Neural Information Processing Systems (NeurIPS)}, 2024.

\bibitem{blindal}
A.-T. Ardelean and T. Weyrich, "Blind Localization and Clustering of Anomalies in Textures," \emph{Proceedings of the IEEE/CVF Conference on Computer Vision and Pattern Recognition Workshops}, 2024.

\bibitem{cola}
A. Saeed, D. Grangier, and N. Zeghidour, "Contrastive Learning of General-Purpose Audio Representations," \emph{arXiv preprint}, 2020.


\bibitem{tripletloss}
F. Schroff, D. Kalenichenko, and J. Philbin, "FaceNet: A Unified Embedding for Face Recognition and Clustering," \emph{Proceedings of the IEEE Conference on Computer Vision and Pattern Recognition (CVPR)}, 2015.


\bibitem{arcface}
J. Deng, J. Guo, N. Xue, and S. Zafeiriou, "ArcFace: Additive Angular Margin Loss for Deep Face Recognition," \emph{2019 IEEE/CVF Conference on Computer Vision and Pattern Recognition (CVPR)}, 2019.

\bibitem{ntxent}
K. Sohn, "Improved Deep Metric Learning with Multi-class N-pair Loss Objective," \emph{Advances in Neural Information Processing Systems (NeurIPS)}, vol. 29, 2016.

\bibitem{infonce}
A. van den Oord, Y. Li, and O. Vinyals, "Representation Learning with Contrastive Predictive Coding," \emph{arXiv}, 2019.

\bibitem{contrastivelift}
Y. Bhalgat, I. Laina, J. F. Henriques, A. Zisserman and A. Vedaldi, "Contrastive Lift: 3D Object Instance Segmentation by Slow-Fast Contrastive Fusion," \emph{Proceedings of the Thirty-seventh Conference on Neural Information Processing Systems}, 2023.

% Related work - Neural Panoptic field
% NeRF
\bibitem{nerf}
B. Mildenhall, P. P. Srinivasan, M. Tancik, J. T. Barron, R. Ramamoorthi and R. Ng, "NeRF: Representing Scenes as Neural Radiance Fields for View Synthesis," \emph{Commun. ACM}, vol. 65, pp. 99–106, Jan. 2022.

\bibitem{moco}
Kaiming He, Haoqi Fan, Yuxin Wu, Saining Xie, and Ross Girshick. "Momentum contrast for unsupervised visual representation learning." \emph{Proceedings of the IEEE/CVF Conference on Computer Vision and Pattern Recognition (CVPR)}, 2020.

\bibitem{zipnerf}
J. T. Barron, B. Mildenhall, D. Verbin, P. P. Srinivasan and P. Hedman, "Zip-NeRF: Anti-Aliased Grid-Based Neural Radiance Fields,"  \emph{Proceedings of the IEEE/CVF International Conference on Computer Vision (ICCV)}, 2023.

\bibitem{kilonerf}
C. Reiser, S. Peng, Y. Liao and A. Geiger, "KiloNeRF: Speeding up Neural Radiance Fields with Thousands of Tiny MLPs," \emph{CoRR}, vol. abs/2103.13744, 2021.

\bibitem{plenoctrees}
A. Yu, R. Li, M. Tancik, H. Li, R. Ng and A. Kanazawa, "PlenOctrees for Real-time Rendering of Neural Radiance Fields," \emph{Proceedings of the IEEE/CVF International Conference on Computer Vision (ICCV)}, 2021.

\bibitem{plenoxels}
S. Fridovich-Keil, A. Yu, M. Tancik, Q. Chen, B. Recht and A. Kanazawa, "Plenoxels: Radiance Fields without Neural Networks," \emph{Proceedings of the IEEE/CVF Conference on Computer Vision and Pattern Recognition (CVPR)}, 2022.

\bibitem{instantngp}
T. Müller, A. Evans, C. Schied and A. Keller, "Instant Neural Graphics Primitives with a Multiresolution Hash Encoding," \emph{ACM Transactions on Graphics (TOG)}, vol. 41, no. 4, pp. 102:1--102:15, July 2022.

\bibitem{zhi2021inplace}
S. Zhi, T. Laidlow, S. Leutenegger and A. J. Davison, "In-Place Scene Labelling and Understanding with Implicit Scene Representation," \emph{CoRR}, 2021.

\bibitem{instancenerf}
Y. Liu, B. Hu, J. Huang, Y.-W. Tai and C.-K. Tang, "Instance Neural Radiance Field," \emph{Proceedings of the IEEE/CVF International Conference on Computer Vision (ICCV)}, 2023.

\bibitem{panopticlifting}
Y. Siddiqui, L. Porzi, S. R. Bulò, N. Müller, M. Nießner, A. Dai and P. Kontschieder, "Panoptic Lifting for 3D Scene Understanding With Neural Fields," \emph{Proceedings of the IEEE/CVF Conference on Computer Vision and Pattern Recognition (CVPR)}, 2023.

% FruitNeRF++ - data preparation

\bibitem{fruitassets}
XFrog Inc., LIBRARY: FRUIT TREES, 2020. [Online: \url{https://www.xfrog.com/product-page/library-fruit-trees}]

\bibitem{BlenderNeRF}
M. Raafat, "BlenderNeRF" (Version 5.0.0), 2023, [Computer software]. \href{https://doi.org/10.5281/zenodo.7926211}{\textcolor{black}{https://doi.org/10.5281/zenodo.7926211}}

\bibitem{fujiapple}
J. Gené-Mola, R. Sanz-Cortiella, J. R. Rosell-Polo, J. R. Morros, J. Ruiz-Hidalgo, V. Vilaplana and E. Gregorio, "Fuji-SfM dataset: A collection of annotated images and point clouds for Fuji apple detection and location using structure-from-motion photogrammetry," \emph{Data in Brief}, vol. 30, 2020.

\bibitem{schoenberger2016sfm}
J. L. Schönberger and J.-M. Frahm, "Structure-from-Motion Revisited," \emph{2016 IEEE Conference on Computer Vision and Pattern Recognition (CVPR)}, 2016.

\bibitem{ren2024grounded}
T. Ren \emph{et al.}, "Grounded SAM: Assembling Open-World Models for Diverse Visual Tasks," \emph{arXiv}, 2024.

\bibitem{liu2023grounding}
S. Liu \emph{et al.}, "Grounding DINO: Marrying DINO with Grounded Pre-Training for Open-Set Object Detection," \emph{arXiv}, 2024.

\bibitem{kirillov2023segany}
A. Kirillov, E. Mintun, N. Ravi, H. Mao, C. Rolland, L. Gustafson, T. Xiao, S. Whitehead, A. C. Berg, W.-Y. Lo, P. Dollár and R. Girshick, "Segment Anything," \emph{arXiv}, 2023.

\bibitem{samhq}
L. Ke, M. Ye, M. Danelljan, Y. Liu, Y.-W. Tai, C.-K. Tang, and F. Yu, "Segment Anything in High Quality," \emph{Proceedings of the Neural Information Processing Systems (NeurIPS)}, 2023.

\bibitem{detic}
X. Zhou, R. Girdhar, A. Joulin, P. Krähenbühl and I. Misra, "Detecting Twenty-thousand Classes using Image-level Supervision," \emph{arXiv}, 2022.

\bibitem{lvis}
A. Gupta, P. Dollár and R. Girshick, "LVIS: A Dataset for Large Vocabulary Instance Segmentation," \emph{arXiv}, 2019.

\bibitem{CLIP}
A. Radford, J. W. Kim, C. Hallacy, A. Ramesh, G. Goh, S. Agarwal, G. Sastry, A. Askell, P. Mishkin, J. Clark, G. Krueger, and I. Sutskever,  
\emph{International Conference on Machine Learning (ICML)}, 2021.  

% FruitNeRF++ -  loss function

\bibitem{thermalnerf}
M. Özer, M. Weiherer, M. Hundhausen, and B. Egger, "Exploring Multi-modal Neural Scene Representations With Applications on Thermal Imaging," \emph{ArXiv}, 2024.

\bibitem{nerfstudio}
M. Tancik, E. Weber, E. Ng, R. Li, B. Yi, J. Kerr, T. Wang, A. Kristoffersen, J. Austin, K. Salahi, A. Ahuja, D. McAllister, and A. Kanazawa, "Nerfstudio: A Modular Framework for Neural Radiance Field Development," \emph{ACM SIGGRAPH 2023 Conference}, 2023.

\bibitem{dbscan}
M. Ester, H.-P. Kriegel, J. Sander and X. Xu, "A Density-Based Algorithm for Discovering Clusters in Large Spatial Databases with Noise," in \emph{Knowledge Discovery and Data Mining}, 1996.

\bibitem{scikitlearn}
F. Pedregosa, G. Varoquaux, A. Gramfort, V. Michel, B. Thirion, O. Grisel, M. Blondel, P. Prettenhofer, R. Weiss, V. Dubourg, J. Vanderplas, A. Passos, D. Cournapeau, M. Brucher, M. Perrot and É. Duchesnay, "Scikit-learn: Machine Learning in Python," \emph{J. Mach. Learn. Res.}, vol. 12, pp. 2825–2830, 2011.

\bibitem{hdbscan}
R. J. G. B. Campello, D. Moulavi and J. Sander, "Density-Based Clustering Based on Hierarchical Density Estimates," in \emph{Advances in Knowledge Discovery and Data Mining}, vol. 7819, pp. 160–172, 2013.

% Evaluation

\bibitem{cl_behaviour}
F. Wang and H. Liu, "Understanding the Behaviour of Contrastive Loss," \emph{arXiv}, 2021. [Online]. Available: https://arxiv.org/abs/2012.09740.

\bibitem{blenderslam}
A. Kalisz, \emph{et al.}, “B-SLAM-SIM: A Novel Approach to Evaluate the Fusion of Visual SLAM and GPS by Example of Direct Sparse Odometry and Blender,” VISIGRAPP, 2019.

% Outlook
\bibitem{gssplatting}
B. Kerbl, G. Kopanas, T. Leimkähler, and G. Drettakis, "3D Gaussian Splatting for Real-Time Radiance Field Rendering," \emph{ACM Transactions on Graphics}, vol. 42, July 2023.

\bibitem{Xu_2023_CVPR}
J. Xu, H. Le, V. Nguyen, V. Ranjan, and D. Samaras, n"Zero-Shot Object Counting," \emph{Proceedings of the IEEE/CVF Conference on Computer Vision and Pattern Recognition (CVPR)}, pp. 15548–15557, 2023.



\if false

\bibitem{POSeg}
A. Kirillov, K. He, R. B. Girshick, C. Rother and P. Dollár, "Panoptic Segmentation," in \emph{2019 IEEE/CVF Conference on Computer Vision and Pattern Recognition (CVPR)}, 2018.


%\bibitem{glomap}
%Linfei Pan, Daniel Barath, Marc Pollefeys, and Johannes Lutz Sch\"{o}nberger, "Global structure-from-motion revisited," European Conference on Computer Vision (ECCV), 2024.



\bibitem{Dong2018SemanticMF}
W. Dong, P. Roy, and V. Isler, "Semantic mapping for orchard environments by merging two-sides reconstructions of tree rows," \emph{Journal of Field Robotics}, vol. 37, pp. 97-121, 2020.doi.org/10.1002/rob.21876).

\bibitem{ICP}
K. S. Arun, T. S. Huang, and S. D. Blostein, "Least-squares fitting of two 3-D point sets," \emph{IEEE Transactions on Pattern Analysis and Machine Intelligence}, 1987.

\fi 


\end{thebibliography}
\end{document}